\documentclass[letter,12pt]{article}
\usepackage[margin=25mm]{geometry}
\usepackage{amsmath}
\usepackage{amsfonts}
\usepackage{amssymb}
\usepackage{graphicx}
\usepackage{verbatim}
\immediate\write18{texcount -tex -sum  \jobname.tex > \jobname.wordcount.tex}

\providecommand{\keywords}[1]
{
  \small	
  \textbf{\textit{Keywords---}} #1
}

\title{Survey on reinforcement learning for language processing}
\author{V\'ictor Uc-Cetina$^{1}$, Nicol\'as Navarro-Guerrero$^{2}$, Anabel Martin-Gonzalez$^{1}$, \\
Cornelius Weber$^{3}$, Stefan Wermter$^{3}$  \\
        \small $^{1}$ Universidad Aut\'onoma de Yucat\'an - \small \{uccetina, amarting\}@correo.uady.mx \\
        \small $^{2}$ Deutsches Forschungszentrum für Künstliche Intelligenz GmbH -
        \small nicolas.navarro@dfki.de \\
        \small $^{3}$ Universit\"at Hamburg -
        \small \{weber, wermter\}@informatik.uni-hamburg.de \\
}
\date{March 2022}

\begin{document}

\maketitle
\begin{abstract}
In recent years some researchers have explored the use of reinforcement learning (RL) algorithms as key components in the solution of various natural language processing tasks. For instance, some of these algorithms leveraging deep neural learning have found their way into conversational systems. This paper reviews the state of the art of RL methods for their possible use for different problems of natural language processing, focusing primarily on conversational systems, mainly due to their growing relevance. We provide detailed descriptions of the problems as well as discussions of why RL is well-suited to solve them. Also, we analyze the advantages and limitations of these methods. Finally, we elaborate on promising research directions in natural language processing that might benefit from reinforcement learning.
\end{abstract}


\section{Introduction}
\label{intro}
Machine learning algorithms have been very successful to solve problems in the natural language processing (NLP) domain for many years, especially supervised and unsupervised methods. However, this is not the case with reinforcement learning (RL), which is somewhat surprising since in other domains, reinforcement learning methods have experienced an increased level of success with some impressive results, for instance in board games such as AlphaGo Zero \cite{silver2017}. Yet, deep reinforcement learning for natural language processing is still in its infancy when compared to supervised learning \cite{LeCun2015Deepa}. Thus, the main goal of this article is to provide a review of applications of reinforcement learning to NLP. Moreover, we present an analysis of the underlying structure of the problems that make them viable to be treated entirely or partially as RL problems, intended as an aid to newcomers to the field. We also analyze some existing research gaps and provide a list of promising research directions in which natural language systems might benefit from reinforcement learning algorithms.

\subsection{Reinforcement learning}
\label{sec:1}
Reinforcement learning is a term commonly used to refer to a family of algorithms designed to solve problems in which a sequence of decisions is needed. Reinforcement learning has also been defined more as a kind of learning problem than as a group of algorithms used to solve such problems \cite{Sutton2018}. It is important to mention that reinforcement learning is a very different kind of learning than the ones studied in supervised and unsupervised methods. This kind of learning requires the learning system, also known as agent, to discover by itself through the interaction with its environment, which sequence of actions is the best to accomplish its goal. 

There are three major groups of reinforcement methods, namely, dynamic programming, Monte Carlo methods, and temporal difference methods. Dynamic programming methods estimate state or state-action values by making estimates from other estimates. This iteratively  intertwines policy evaluation and policy improvement updates taking advantage of a model of the environment which is used to calculate rewards. Policy evaluation consists of updating the current version of the value function based on the current policy. Policy improvement consists of greedifying the policy function based on the current value function. Depending on the algorithm and its implementation it might require exhaustive sweeping of the entire state space or not. Monte Carlo methods learn from complete sample returns, instead of immediate rewards. Unlike dynamic programming, Monte Carlo methods only consider one transition path at a time, the path generated with a sample. In other words, they do not bootstrap from successor states’ values. Therefore, these kinds of methods are more useful when we do not have a model of the environment, the so-called dynamics of the environment. Temporal difference methods do not need a model of the environment since they can learn from experience, which can be generated from interactions with the environment. These methods possess the best of dynamic programming and the best of Monte Carlo. From dynamic programming they inherit the bootstrapping, from Monte Carlo methods they inherit the sampling. As a result of this combination of characteristics, temporal difference methods have been the most widely used. All these methods pose the decision-making problem as a Markov decision process (MDP). An MDP is a mathematical method used to solve decision-making in sequence and considers as the minimum existing elements a set of states $S$, a set of actions $A$, a transition function $T$, and a reward function $R$. Given an MDP $(S, A, T, R)$, we need to find an optimal policy function $\pi$, which represents the solution of our sequence decision problem. The aim of a reinforcement learning system, or so-called agent, is to maximize some cumulative reward  $r \in R$ through a sequence of actions. Each pair of state $s$ and action $a$ creates a transition tuple $(s,a,r,s')$, with $s'$ being the resulting state. Depending on the algorithm being used and on the particular settings of our problem, the policy $\pi$ will be estimated differently.

A policy $\pi$ defines the behavior of the agent at any given moment. In other words, a policy is a mapping from the set of states $S$ perceived from the environment to a set of actions $A$ that should be executed in those states. In some cases, the policy can be stored as a lookup table, and in other cases it is stored as a function approximator, such as a neural network. The latter is imperative when we have a large number of states. The policy is the most important mathematical function learned by the reinforcement learning agent, in the sense that it is all the agent needs to control its behavior once the learning process has concluded. In general, a policy can be stochastic and we formally define it as $\pi: S \rightarrow A$.

The goal in a reinforcement learning problem is specified by the reward function $R$. This function maps each state or pair of state-action perceived in the environment to one real number $r \in \Re$ called reward. This reward indicates how good or bad a given state is. As we mentioned before, the goal of an agent is to maximize the total amount of rewards that it gets in the long run, during its interaction with the environment. The reward function cannot be modified by the agent, however, it can serve as a basis for modifying the policy function. For example, if the action selected by the current policy is followed by a low reward, then the policy can be updated in such a way that in the future it indicates a different action when the agent encounters the same situation. In general, the reward function can also be a stochastic function and it is formally defined as $R: S \rightarrow \Re$, or $R: S \times A \rightarrow \Re$.

A value function indicates which actions are good in the long run. The value of a state is basically an estimation of the total amount of rewards that the agent can expect to accumulate in the future, if it starts its path from that state using its current policy. We should not confuse the value function with the reward function. The rewards are given directly by the environment while the values of the states are estimated by the agent, from its interaction with the environment. Many reinforcement learning methods estimate the policy function from the value function. When the value function is a mapping from states to real numbers, it is denoted by the letter $V$. When the mapping is from pairs of state-action to real numbers, it is denoted by $Q$. Formally, we can define the value function as $V: S \rightarrow \Re$ or $Q: S \times A\rightarrow \Re$.

In the case of model-based RL, the agent also has access to a model of the transition function $T$ of the environment, which may be learnt from experience.
For example, given a state and an action, the model could predict the next resulting state and reward. Such world models are used for planning, this is, a way to make decisions about the next actions to be performed, without the need to experience possible situations. In the case of model-free RL, when a model of the environment is missing, we have to solve the reinforcement learning problem without planning and that means that a significant amount of experimentation with the environment will be needed.

One of the most popular reinforcement learning algorithms is the Q-learning algorithm \cite{Watkins1989}. As its name suggests, it works by estimating a state-action value function $Q$. The algorithm does not rely on a model of the transition function $T$, and therefore it has to interact with the environment iteratively. It follows one policy function for exploring the environment and a second greedy policy for updating its estimations of the values of pairs of states and actions that it happens to visit during the learning process. This kind of learning is called off-policy learning. The algorithm uses the following rule for updating the $Q$ values:
\[Q(s,a) \leftarrow Q(s,a) + \alpha [r + \gamma \max_{a'} Q(s',a') -  Q(s,a)].\]

In this learning rule, $\alpha$ is a parameter defined experimentally and it is known as the learning rate. It takes values in the interval $(0, 1)$. Moreover, $r$ is the reward signal, $\gamma$ is known as the discount parameter and it also takes values in the interval $(0,1)$, and finally $s'$ and $a'$ denote the next state and the next action to be visited and executed during the next interaction with the environment. 

SARSA is an on-policy learning algorithm, meaning that instead of using two policies, one for behavior and one for learning, this algorithm uses only one policy. The same policy that is used to explore the environment is the same policy used in the update rule \cite{Sutton2018}. The update rule of the SARSA is the following:
\[Q(s,a) \leftarrow Q(s,a) + \alpha [r + \gamma Q(s',a') - Q(s,a)].\]

A very important result in recent years was the development of the deep Q-network \cite{Mnih2015}, in which a convolutional neural network is trained with a variant of Q-learning. This algorithm, originally designed to learn to play several Atari 2600 games at a superhuman level, is now being applied to other learning tasks. Another algorithm, AlphaGo Zero \cite{Silver2016},  learned to play Go and actually defeated the human world champion in 2016. This algorithm uses a deep neural network, a search algorithm and reinforcement learning rules. The successor model MuZero \cite{Schrittwieser2019} learns a representation of state, a dynamics and a reward prediction function to maximize future rewards via tree search-based planning, achieving more successful game play without prior knowledge of the game rules. 

Deep reinforcement learning is an extension of the classical reinforcement learning methods to leverage the representational power of deep models. More specifically, deep neural networks allow reinforcement learning algorithms to approximate and store highly complex value functions, state-action functions, or policy functions. For instance, a $Q(s,a)$ function can be represented as a convolutional neural network or a recurrent one. Similarly to what happened in other domains such as computer vision, deep models are also playing a decisive role in the advancement of reinforcement learning research, especially in MDPs with very large state and action spaces. In fact, reinforcement learning and deep neural networks have stayed recently at the center of attention of many researchers who have studied and applied them to solve different problems, including problems in natural language processing, as we will discuss below.

\subsection{Natural language processing and RL}
\label{sec:2}
In natural language processing, one of the main goals is the development of computer programs capable of communicating with humans through the use of natural language. In some applications, such as machine translation, these programs are used to help humans who speak different languages to understand each other by translating from one natural language to another. Through the years, NLP research has gone from being heavily influenced by theories of linguistics, such as those proposed by Noam Chomsky~\cite{chomsky1959certain,chomsky65UG}, to the corpus linguistics approach of machine learning algorithms and more recently the use of deep neural networks as neural language models such as BERT \cite{Devlin2018} and GPT-3 \cite{brown2020language}. 
 
According to Russell and Norvig \cite{Russell2009}, to the contrary of formal languages, it is more fruitful to define natural language models as probability distributions over sentences rather than using definitive sets specified by grammars. The main challenges when dealing with natural languages are that they are ambiguous, large and constantly changing. That is why initial approaches to model natural languages using grammars were not as successful as modern machine learning approaches. In the former approaches, the grammars needed to be adapted and their size increased to fulfil the demands for better performance.

One important probabilistic approach to modelling natural languages involves the use of $n$-grams. A sequence of written symbols of length $n$ is called an $n$-gram. A model of the probability distribution of strings containing $n$ symbols is therefore called an $n$-gram model. This model is defined as a Markov chain of order $n-1$ in which the probability of some symbol $s_i$ depends only on the immediately preceding $n-1$ symbols. Formally, we say $p(s_i | s_{i-1}, s_{i-2}, \ldots, s_{2}, s_{1}) =  p(s_i | s_{i-1}, \ldots, s_{i-n+1})$. Probabilistic natural language models based on $n$-grams can be useful for text classification tasks \cite{Russell2009}.

 \begin{figure}[ht]
 \centering
 \includegraphics[width=11cm]{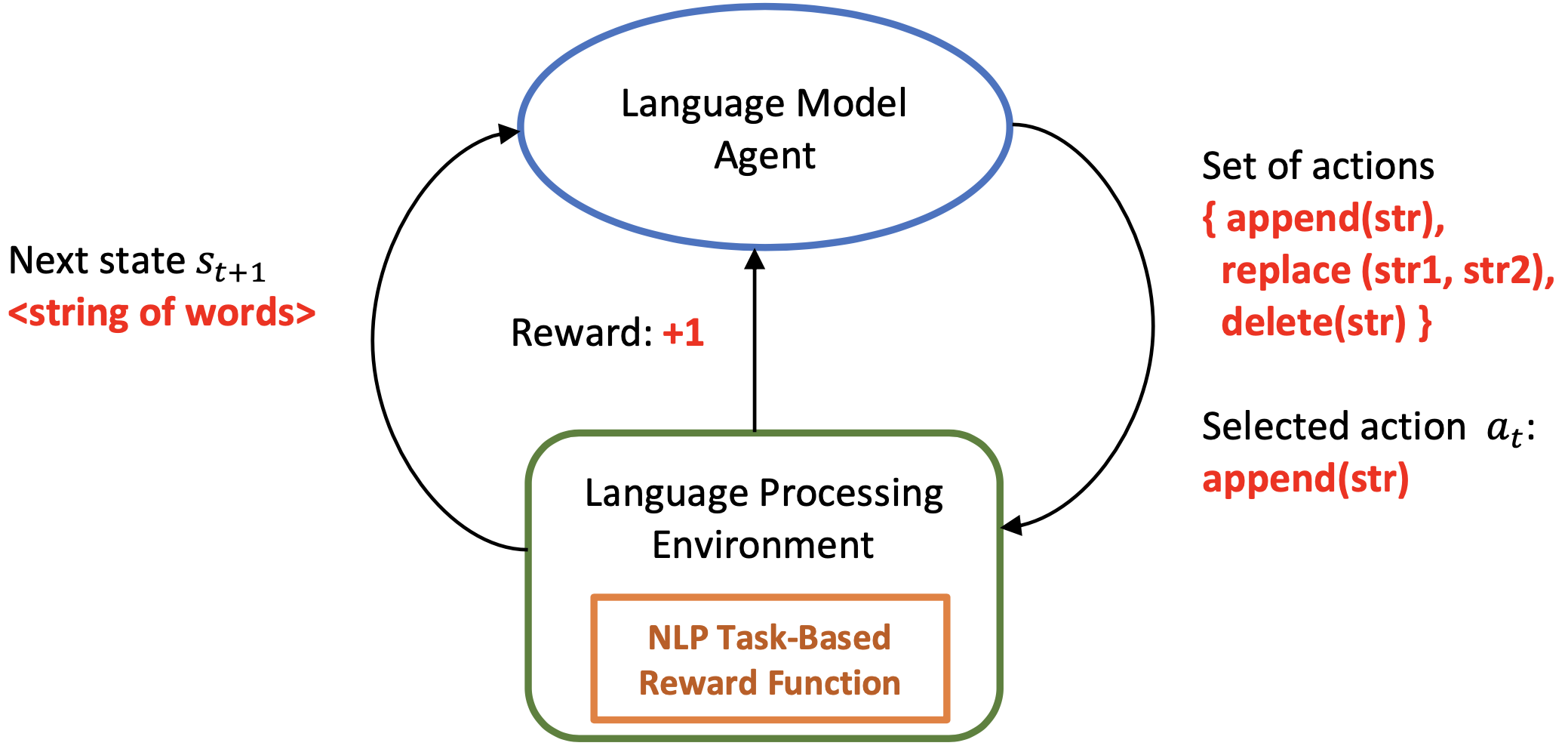}
 \caption{Schematic view of a reinforcement learning agent designed for language processing. The language model agent acts by appending, replacing or deleting strings of words. States are strings of words. The language processing environment will provide the agent with the states and rewards after each of the interactions. The reward function is determined by the specific natural language processing task. One simple possibility for a reward function would reinforce every optimal action with a +1.}
 \label{rl_for_nlp}
 \end{figure}

Important advances in the design of algorithms for training deep neural networks, such as the recurrent long short-term memory (LSTM) network \cite{Hochreiter1997}, have allowed researchers to move from probabilistic language models to language models based on neural networks. The LSTM neural model has been successfully applied to machine translation. The performance of current translator programs could not be accomplished using the approach based on language grammars alone. These new neural models are highly complex mathematical functions with thousands of parameters which are estimated iteratively from a massive number of training examples gathered from the Internet.

Some problems in natural language processing can be defined as Markov decision processes and therefore they can be solved using reinforcement learning algorithms. In Fig.\ \ref{rl_for_nlp}, we provide a schematic example of how a reinforcement learning agent would be designed to solve a language processing task in which states, actions and rewards operate mainly over strings. A set of basic operations may include appending, replacing and deleting words.

In this article, we review five main categories of such problems, namely,  syntactic parsing, language understanding, text generation systems, machine translation, and conversational systems. Of these, conversational systems are the most studied ones, which involve finding an optimal dialog policy that should be followed by an automated system during a conversation with a human user. The other four categories are not widely known applications of reinforcement learning methods and therefore it is interesting to discuss their main benefits and drawbacks. In some of them, it is even not easy to identify the elements of a well-defined Markov decision process. This might explain why they have not received more attention yet. Identifying these different natural language processing problems is important to discover new research lines at the intersection of natural language processing and reinforcement learning.

In the next sections, we describe with more detail these five categories of natural language processing problems and their proposed solutions by means of reinforcement learning. We also discuss the main achievements and core challenges on each of these categories.

\section{Syntactic parsing}
\label{parsing}
Syntactic parsing consists of analyzing a string made of symbols belonging to some alphabet, either in natural languages or in programming languages. Such analysis is often performed according to a set of rules called grammar. There could be many ways to perform parsing, depending on the final goal of the system \cite{zhang2009,jiang2012,neu2009,le2017}. One of such goals could be the construction of a compiler for a new programming language when we are working with formal computer languages. Another one could be an application of language understanding for human-computer interaction.

A grammar can generate many parsing trees and each of these trees specifies the valid structure for sentences of the corresponding language. Since parsing can be represented as a sequential search problem with a parse tree as the final goal state, reinforcement learning methods are tools very well suited for the underlying sequential decision problem. In general, a parse is obtained as a path when an optimal policy is used, in a given MDP.

Consider for example the simple context-free grammar $G_1$ and the language $L(G_1)$ generated by it. $G_1$ is a 4-tuple $(V, \Sigma, R, S)$ where

\begin{itemize}
 \item $V = \{ A, B\}$ is a finite set of variables,
 \item $\Sigma = \{0,1,\#\}$ is a finite set, disjoint of $V$, containing terminal symbols,
\item $R$ is the finite set of 4 production rules given in Fig.\ \ref{grammar01}, and
\item $S \in V$ is the initial variable.
\end{itemize}

\begin{figure}[h!]
 \centering
\begin{align}
 S &\rightarrow 0A1 \\
 A &\rightarrow 0A1 \; | \;  B\\
 B &\rightarrow \#
\end{align}
\caption{Grammar $G_1$ with 4 production rules.}
\label{grammar01}
\end{figure}

\begin{figure}[ht]
 \centering
 \includegraphics[width=4cm]{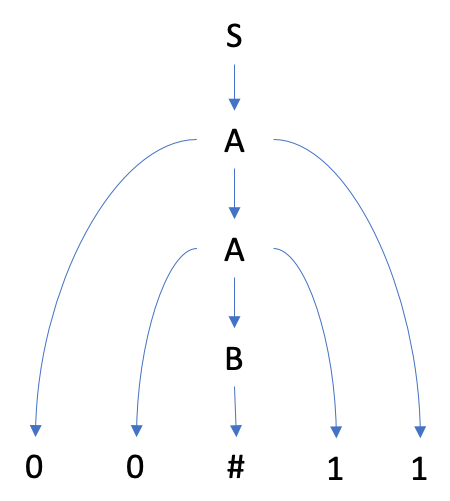}
 \caption{Parse tree of string $00\#11$ generated from grammar $G_1$.}
 \label{ptree01}
 \end{figure}

The language $L(G_1)$ generated by grammar $G_1$ is an infinite set of strings. Each of these strings is created by starting with the initial variable $S$ and iteratively selecting and applying one of the production rules in $G_1$, also called substitution rules. For example, the string $0\#1$ is a valid string belonging to $L(G_1)$ and it can be generated by applying the following sequence of production rules $S \rightarrow 0A1$, $A \rightarrow B$ and $B \rightarrow \#$. Looking at this application of rules as a path of string substitutions, we have $S \Rightarrow 0A1 \Rightarrow 0B1 \Rightarrow 0\#1$. A path of substitutions, known also as derivation, can be represented pictorially as a parse tree. For example, the parse tree for the derivation of the string $00\#11$ is illustrated in Fig.\ \ref{ptree01}.

 From the previous grammar example $G_1$ we can notice the similarity between the elements defined in a context-free grammar $ G = \{V, \Sigma, P, S\}$ and the elements defined in a Markov decision process $M = \{S,A,T,R\}$. Let us now analyze this similarity, element by element, from the point of view of an MDP. 

 \begin{itemize}
\item The starting state $s$ of an MDP $M$ can be defined as the initial variable of a grammar, denoted by letter $S$ in grammar $G$.

\item The set of states $S$ in the MDP $M$ can be defined as the set of strings generated by the grammar, in other words, the language generated by grammar $G$, this is $S = L(G)$.

\item The set of actions $A$ can be defined as the set of production rules given by grammar $G$, this is $A=R$; the MDP transition function $T$ would be immediately defined once we have defined the set of production rules itself.

\item The reward function $R$ is the only element that cannot be taken straightforward from the elements of the grammar and it should be crafted by the designer of the system.
 \end{itemize}
 
In the specific application of dependency parsing \cite{Kubler2009}, it has been shown that a parser can be implemented to use a policy learned by reinforcement learning, in order to select the optimal transition in each parsing stage \cite{zhang2009}. Given a sentence with $n$ words $x = w_1 w_2 \ldots w_n$, we can construct its dependency tree by selecting a sequence of transitions. A stack data structure is used to store partially processed words and also a queue data structure is used to record the remaining input words together with the partially labeled dependency structure constructed by the previous transitions. The construction of the dependency tree is started with an empty stack and the input words being fed into the queue. The algorithm performs 4 different types of transitions until the queue is empty. These 4 transitions are: \textit{reduce}, which takes one word from the stack; \textit{shift}, which pushes the next input word into the stack; \textit{left-arc}, which adds a labeled dependency arc from the next input word to the top of the stack and then takes the word from the top of the stack; and finally \textit{right-arc}, which adds a dependency arc from the top of the stack to the next input word and pushes that same word into the stack. During the construction of the parsing tree each one of the transitions is selected using a reward signal. In this particular implementation the optimal policy for selecting the transitions is estimated using the SARSA reinforcement learning algorithm. 

An interesting modification found in the implementation of this algorithm is the replacement of the $Q$ function by an approximation computed through the calculation of the negative free energies of a restricted Boltzmann machine. The results of this approach for dependency parsing using reinforcement learning are comparable with state-of-the-art methods. More recently, it has been shown that reinforcement learning can also be used to reduce error propagation in greedy dependency parsing \cite{le2017}. In another approach, Neu et al.\ \cite{neu2009} used a number of inverse reinforcement learning algorithms to solve the parsing problem with probabilistic context-free grammars. In inverse reinforcement learning, given a set of trajectories in the environment, the goal is to find a reward function such that if it is used for estimating the optimal policy, the resulting policy can generate trajectories very similar to the original ones \cite{Ng2000}. 

Another dual learning approach for solving the semantic parsing problem is presented by Cao et al.\ \cite{CaoZhu2019}. This dual learning algorithm follows the same strategy used by Zhu et al.\ \cite{ZhuCao2020}, consisting of an adversarial training scheme that can use both labeled and unlabeled data. The primary task (semantic parsing) learns the transformation from a query to logical form (Q2LF). The secondary task (natural language generation) learns the transformation from a logical form to a query (LF2Q). The agent from the primary task can teach the agent from the secondary task and vice versa in a reinforcement learning fashion. A validity reward by checking the output of the primary model at the surface and at semantic levels is used. This reward function requires prior knowledge of the logical forms of the domain of interest, and it is used to check for completeness and well-formed semantic representations. The experimental results showed that semantic parsing based on dual learning improves performance across datasets.

In a probabilistic context-free grammar, each production rule has a probability assigned to it, which results in the generation of expert trajectories. Speeding up the learning of parse trees using reinforcement learning has also been studied, specifically the use of apprenticeship reinforcement learning as a variation of inverse RL has been shown to be an effective method for learning a fast and accurate parser, requiring only a simple set of features \cite{jiang2012}. 
By abstracting the core problem in syntactic parsing, we  can clearly see that it can be posed as an optimization problem in which the input is a language grammar $G$ and one input string $w_1$ to be parsed, and the output is a parse tree that allows the correct parsing of $w_1$.  This problem gives rise to the following MDP $(S,A,T,R)$  \cite{le2017}:

\begin{itemize}
\item The set of states $S$ is defined as the set of all possible partial or complete parse trees that can be generated with the given grammar $G$ and the string $w_1$.
\item The set of actions $A$ is formed with all the grammar rules contained in $G$, this is, the application of each derivation rule of the grammar is considered to be an action. 
\item The transition function $T$ can be completely determined and it is deterministic, because given a selected grammar rule and the current partially parsed string, we can know with certainty the next resulting intermediate parse tree of that string. 
\item Finally, the reward function $R$ can be defined as a function of the number of arcs that are correctly labeled in the resulting parse tree. 
\end{itemize}

\begin{figure}[ht]
 \centering
 \includegraphics[width=11cm]{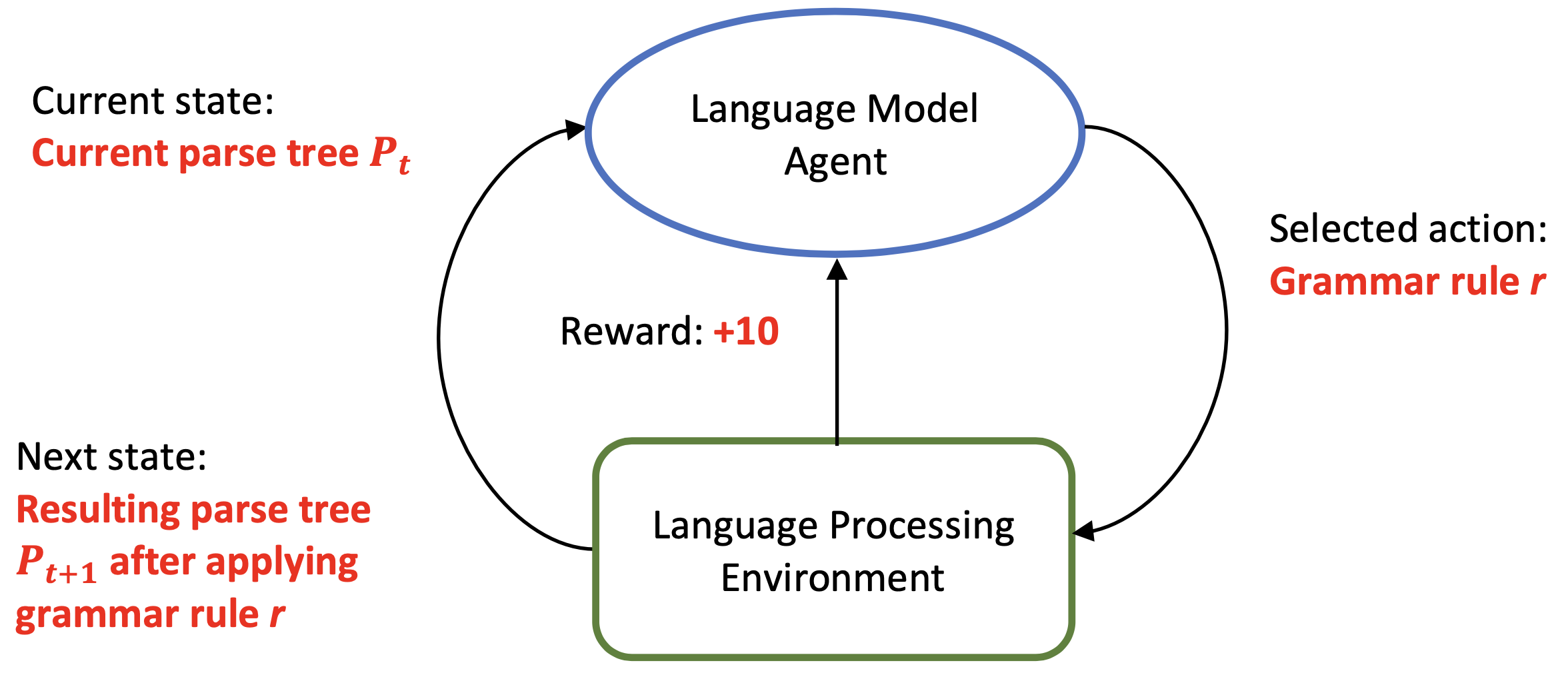}
\caption{Schematic view of a reinforcement learning agent designed for syntactic parsing. The language processing environment will provide the agent with the states and rewards after each of the interactions. The reward function can be defined in various ways, for example, a positive reward of 10 may be provided each time an appropriate grammar rule is applied.}
 \label{parsing_scheme}
 \end{figure}

Based on this MDP we can formulate a reinforcement learning system as illustrated in Fig.\ \ref{parsing_scheme}.

\section{Language understanding}
Language understanding can also be posed as a Markov decision process and therefore we can apply sophisticated reinforcement learning algorithms designed in recent years. Furthermore, we can implement them together with deep neural networks to cope with the massive amount of data that text understanding applications typically require. 

\begin{figure}[h!]
\begin{align*}
 \langle\mathtt{SENTENCE}\rangle &\rightarrow \langle\mathtt{NOUN\_PHRASE}\rangle \langle\mathtt{VERB\_PHRASE}\rangle\\
\langle\mathtt{NOUN\_PHRASE}\rangle &\rightarrow \langle\mathtt{CMPLX\_NOUN}\rangle \; | \;  \langle\mathtt{CMPLX\_NOUN}\rangle \langle\mathtt{PREP\_PHRASE}\rangle \\
 \langle\mathtt{VERB\_PHRASE}\rangle & \rightarrow \langle\mathtt{CMPLX\_VERB}\rangle \; | \;  \langle\mathtt{CMPLX\_VERB}\rangle \langle\mathtt{PREP\_PHRASE}\rangle \\
 \langle\mathtt{PREP\_PHRASE}\rangle & \rightarrow \langle\mathtt{PREP}\rangle \langle\mathtt{CMPLX\_NOUN}\rangle \\
 \langle\mathtt{CMPLX\_NOUN}\rangle & \rightarrow \langle\mathtt{ARTICLE}\rangle \langle\mathtt{NOUN}\rangle \\
 \langle\mathtt{CMPLX\_VERB}\rangle & \rightarrow \langle\mathtt{VERB}\rangle \; | \; \langle\mathtt{VERB}\rangle \langle\mathtt{NOUN\_PHRASE}\rangle \\
 \langle\mathtt{ARTICLE}\rangle & \rightarrow \mathtt{a} \; | \;  \mathtt{the} \\
 \langle\mathtt{NOUN}\rangle & \rightarrow \mathtt{customer} \; | \; \mathtt{discount} \; | \; \mathtt{refund} \\
 \langle\mathtt{VERB}\rangle & \rightarrow \mathtt{wants} \; | \; \mathtt{requests} \; | \; \mathtt{cancelled} \\
 \langle\mathtt{PREP}\rangle & \rightarrow \mathtt{with}
 \end{align*}
\caption{Grammar defining valid sentences in English, Grammar adapted from \cite{Sipser2013}.}
\label{grammar02}
\end{figure}

\begin{figure}[ht]
 \centering
 \includegraphics[width=11cm]{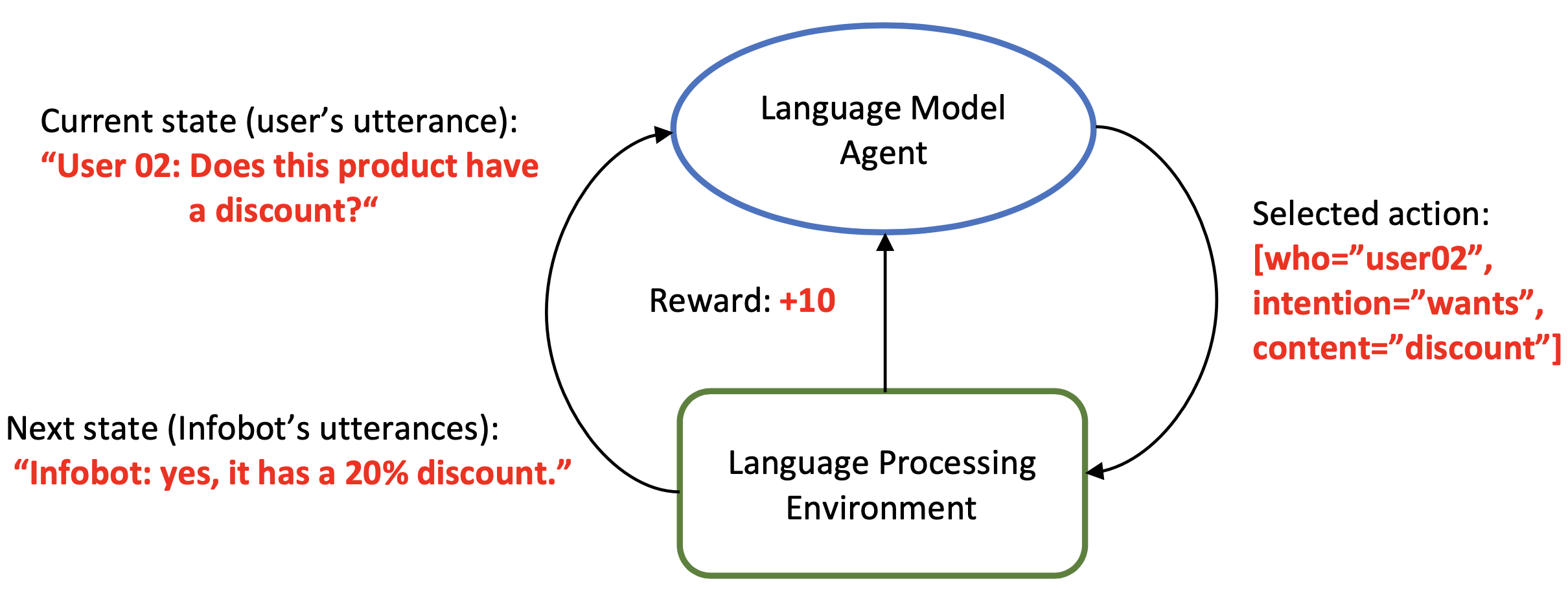}
\caption{Schematic view of a reinforcement learning agent designed for text understanding, as an example application. The language model agent acts by assigning values to a vector or list of variables, as a function of the utterance of a user. The current user $user02$ is computed from the user's utterance. The next state is the string generated from the vector of variables as understood by the agent, for example, using a text generator agent (see Fig.\ \ref{text_generator_agent}). The language processing environment will provide the agent with the states and rewards after each of the interactions. The environment and the reward function are determined by the language understanding  task being solved, i.e., an infobot.}
 \label{understanding_scheme}
 \end{figure}

Consider a problem of natural language understanding. In such a problem we could have a grammar like the one given in Fig.\ \ref{grammar02} that allows a program to automatically determine the elements of a sentence written in English. Using this grammar, sentences such as ``The customer with a discount wants a refund'' and ``The customer with a discount cancelled the refund'' can be analyzed by an automated system to determine the intention of the customer, which in this case is whether she wants a refund or she wants to cancel a refund she had previously requested. Therefore, a grammar can be used to detect users' intentions while reinforcement learning can be used to select the optimal sequence of substitutions during the parsing process of the input sentences. Once the parser program has been used to determine the grammatical role of each word in the input text string, the result can be stored in a vector-type structure such as [\emph{who}=user02, \emph{intention}=``wants'', \emph{content}=``discount'']. This vector-type representation of variables \emph{who}, \emph{intention} and \emph{content}, can be used for another program to determine the most appropriate action to be performed next. For example, informing about a discount to a customer. Figure \ref{understanding_scheme} outlines the procedure.

Ambiguities are an important problem in language understanding. For example, the sentence ``the child observes the cat in the tree'' may have two interpretations, whether the child is in the tree or the cat is in the tree. This kind of ambiguity in the language is hard to solve even by humans. Sometimes it can be solved by using context or common sense. From the point of view of reinforcement learning, there is no obvious way to solve it either. One approach to this problem would be to leverage the powerful text embedding vectors generated by sophisticated language models such as GPT together with a function that rewards making corrections as learning interactions go on, taking advantage of the context. GPT-based models are very good at keeping contextual information. A reward function could provide a larger reward when the interpretation of the intent is more highly evaluated by a context metric provided by the language model.

Language understanding programs approached by reinforcement learning have to deal with systems that automatically interpret text or voice in the context of a complex control application, and use the knowledge extracted to improve control performance. Usually, the text analysis and the learning of the control strategy are carried out both at the same time. For example, Vogel and Jurafsky \cite{vogel2010} implement a system capable to learn to execute navigational instructions expressed in a natural language. The learning process is carried out using an apprenticeship approach, through pairs of paths in a map and their corresponding descriptions in English. The challenge here is to discover which commands match English instructions for navigation. The correspondence is learned applying reinforcement learning and using the deviation between the given desired path and the route being followed for the reward signal. This work demonstrates that the semantic meaning of spatial terms can be grounded into geometric properties of the paths. In a similar approach to language grounding \cite{branavan2012} the system learns to interpret text in the context of a complex control application. Using this approach, text analysis and control strategies are learned jointly using a neural network and a Monte Carlo search algorithm. The approach is tested on a video game, using its official manual as a text guide.

Deep reinforcement learning has also been used to automatically play text games \cite{he2016}, showing that it is possible to extract meaning rather than simply memorizing strings of texts. This is also the case of the work presented by \cite{guo2017}, where an LSTM and a Deep Q-Network are employed to solve the sequence-to-sequence problem. This approach is tested with the problem of rephrasing a natural language sentence. The encoding is performed using the LSTM and the decoding is learned by the DQN. The LSTM initially suggests a list of words which are taken by the DQN to learn an improved rephrasing of the input sentences.

Zhu et al.\ \cite{ZhuCao2020} presented a semi-supervised approach to tackle the dual task of intent detection and slot filling in natural language understanding (NLU). The suggested architecture consists of a dual pseudo-labeling method and a dual learning algorithm. They apply the dual learning method by jointly training the NLU and semantic-to-sentence generation (SSG) models, using one agent for each model. As the feedback rewards are non-differentiable, a reinforcement learning algorithm based on policy gradient is applied for optimization. The two agents collaborate in two closed loops. The NLU2SSG loop starts from a sentence, first generating a possible semantic form by the NLU agent and then reconstructing the original sentence by SSG. The SSG2NLU loop goes in reverse order. Both the NLU and SSG models are pre-trained on labeled data. The corresponding validity rewards for the NLU and SSG evaluate whether the semantic forms are valid. The approach was evaluated on two public datasets, i.e., ATIS and SNIPS, achieving state-of-the-art performance. The proposed framework is agnostic of the backbone model of the NLU task.

Text understanding is one of the most recent natural language problems approached using reinforcement learning, specifically by deep reinforcement learning. This approach consists of mapping text descriptions into vector representations. The main goal is to capture the semantics of the texts. Therefore, learning good representations is key. In this context, it has been argued that LSTMs are better than Bag-Of-Words (BOW) when combined with reinforcement learning algorithms. The reason is that LSTMs are more robust to small variations of word usage, and they can learn some underlying semantics of the sentences \cite{narasimhan2015}.

As we have seen above, the main applications of reinforcement learning in the context of language understanding have been focused on the learning of navigational directions. RL or inverse RL recommend themselves over supervised learning due to the good match between sequential decision making and parsing. However, it is not difficult to think of other similar applications that could take advantage of this approach. For example, if we can manage to design a system capable to understand text to some degree of accuracy, such a system could be used to implement intelligent tutors, smart enough to understand the questions posed by the user and select the most appropriate learning resource, whether it is some text, audio, video, hyperlink, etc.

Interestingly, the successful results recently obtained with the combination of deep neural networks and reinforcement learning algorithms open another dimension of research that appears to be promising in the context of parsing and text understanding. As we have mentioned before, creating natural language models is difficult because natural languages are large and constantly changing. We think that deep reinforcement learning (DRL) could become the next best approach to natural language parsing and understanding. Our reasoning is based primarily on two facts. First, DRL can store optimally thousands of parameters of the grammars as a neural model, and we have already evidence that these neural models can be very effective with other natural language problems such as machine translation. Second, reinforcement learning methods would allow the agent to keep adapting to changes in a natural language, since the very nature of these algorithms is to learn through interaction and this feature allows the reinforcement learning agents to constantly adapt to changes in their environment.

\section{Text generation systems}
\label{generation}
Text generation systems are built to automatically generate valid sentences in natural language. One of the components of such systems is a language model. Once the language model is provided or learned, the optimization problem consists of generating valid sequences of substrings that will subsequently complete a whole sentence with some meaning in the domain of the application.

Given a vector representation of a set of variables in a computational system and their corresponding values, a reinforcement learning algorithm can be used to generate a sentence in English, or any other natural language, that can serve to communicate specific and meaningful information to a human user. However, using the information stored in a set of program variables and constructing sentences in a natural language representing such information is not an easy task. This problem has been studied in the context of generating navigational instructions for humans, where the first step is to decide about the content that the system wants to communicate to the human, and the second step is to build the correct instructions adding word by word. An interesting point in this approach is that the reward function is implemented as a hidden Markov model \cite{dethlefs2011} or as a Bayesian network \cite{dethlefs2011a}. The reinforcement learning process is carried out with a hierarchical algorithm using semi-MDP's.
 
 \begin{figure}[ht]
  \centering
 \includegraphics[width=11cm]{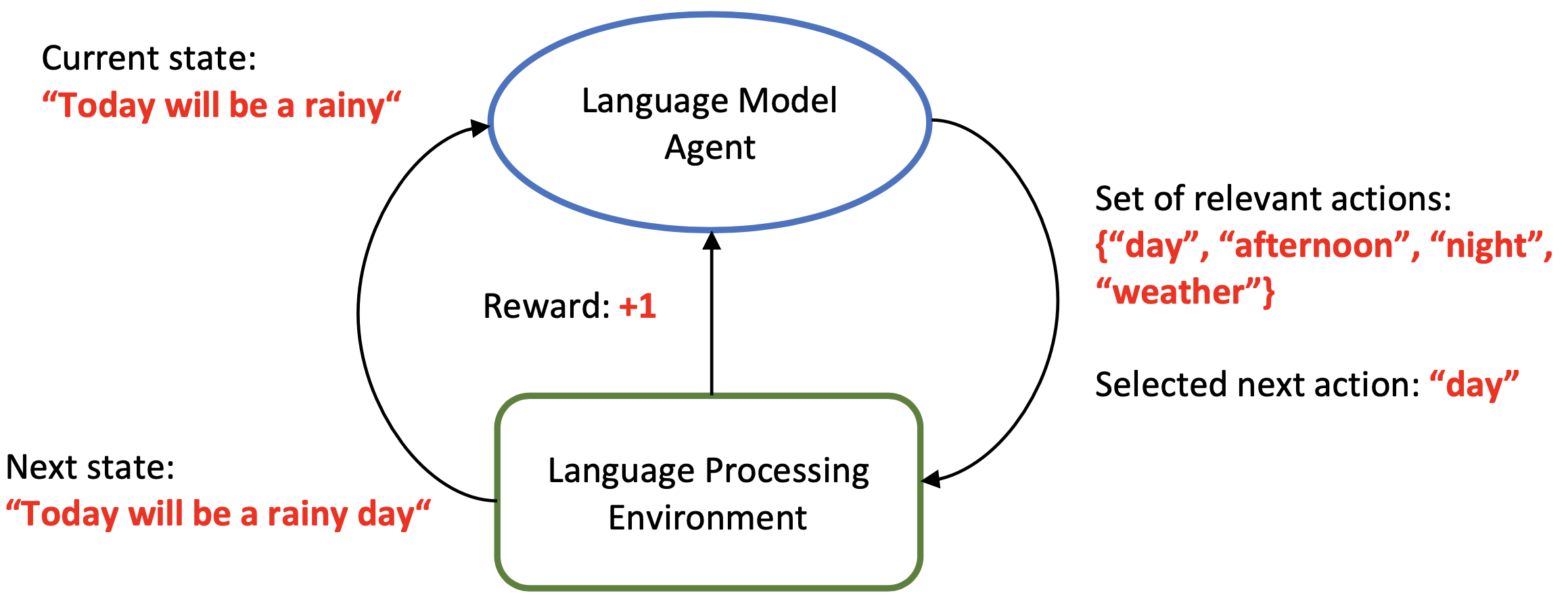}
  \caption{Schematic view of a reinforcement learning agent designed for language generation, as an example application. The language model agent acts by selecting words from a relevant set of words, which is a function of the current state. The current state is a -- possibly incomplete -- sentence in English. The next state is the sentence resulting from appending the word selected by the agent. The language processing environment will provide the agent with the states and rewards after each of the interactions. Actions might take the form of strings of characters such as n-grams, words, sentences, paragraphs or even full documents. The environment and the reward function are determined by the language processing task being solved, i.e., text generation.}
  \label{text_generator_agent}
 \end{figure}

Text generation has also been  approached using inverse reinforcement learning (IRL) \cite{Ziebart2008} and generative adversarial networks (GANs) \cite{Goodfellow2014}. Shi et al.\ \cite{Shi2018} proposed a new method combining GANs and IRL to generate text. The main result of this work is the alleviation of two problems related to generative adversarial models, namely reward sparsity and mode collapse. The authors of this work also introduced new evaluation measures based on BiLingual Evaluation Understudy (BLEU) score, designed to evaluate the quality of the generated texts in terms of matching human-generated expert translations. They showed that the use of IRL can produce more dense reward signals and it can also generate more diversified texts. With this approach, the reward and the policy functions are learned alternately, following an adversarial model strategy. According to the authors, this model can generate texts with higher quality than previous proposed methods based also on GANs, such as SeqGAN \cite{Yu2017-SeqGAN}, RankGAN \cite{Lin2017}, MaliGAN \cite{Che2017} and LeakGAN \cite{Guo2018}. The adversarial text generation model uses  a discriminator and a generator. The discriminator judges whether a text is real or not, meanwhile the generator learns to generate texts by maximizing a reward feedback provided by the discriminator through the use of reinforcement learning. The generation of entire text sequences that these adversarial models can accomplish helps to avoid the exposure bias problem, a known problem experienced by text generation methods based on RNNs. The exposure bias problem \cite{Bengio2015} lets small discrepancies between the training and inference phases accumulate quickly along the generated sequence.

In a text generation task the corresponding MDP might be defined as follows:

\begin{itemize}
\item Each state in $S$ is formed with a feature vector describing the current state of the system being controlled, containing enough information to generate the output string. We can visualize this feature vector as a set of variables that describe the current status of the system.
\item Actions in $A$ will consist of adding or deleting words.
\item With respect to the transition function $T$, every next state can be determined by the resulting string, after we have added or deleted a word.
\item In this task, the reward function could be learned from a corpus of labeled data or more manually, from human feedback.
\end{itemize}

An advantage of RL methods over supervised learning for text generation becomes apparent when there is a diversity of valid text output, i.e., multiple different generations would be of equal quality. In this case, it is problematic for supervised learning to define a differentiable error for backpropagation. However, evaluation measures like BLEU or the Recall-Oriented Understudy for Gisting Evaluation (ROUGE) can be used well to define a reward function for RL \cite{Keneshloo2019}. Future research work can focus on adaptive natural language generation during human-computer interaction, assuming a continuously changing learning environment. In natural language generation the main goal is to build a precise model of the language, and the current existing approaches are far from being generic.

Another more complicated possibility is the study of language evolution under a reinforcement learning perspective. In general, language evolution is concerned with how a group of agents can create their own communication system \cite{Cangelosi2002}. The communication system emerges from the interaction of a set of agents inhabiting a common environment. A process like this can be modeled as a reinforcement learning multi-agent system \cite{Mordatch2018}.

Li et al.\ \cite{Li2018Paraphrase} used reinforcement learning and inverse reinforcement learning for paraphrase generation. One of the components of this approach is a generator. The generator is initially trained using deep learning and then it is fine-tuned using RL. The reward of the generator is given by a second component of the architecture, the evaluator. The evaluator is a deep model trained using inverse RL to evaluate whether two given phrases are similar to each other.

\section{Machine translation}
\label{translation}
\begin{figure}[ht]
 \centering
 \includegraphics[width=11cm]{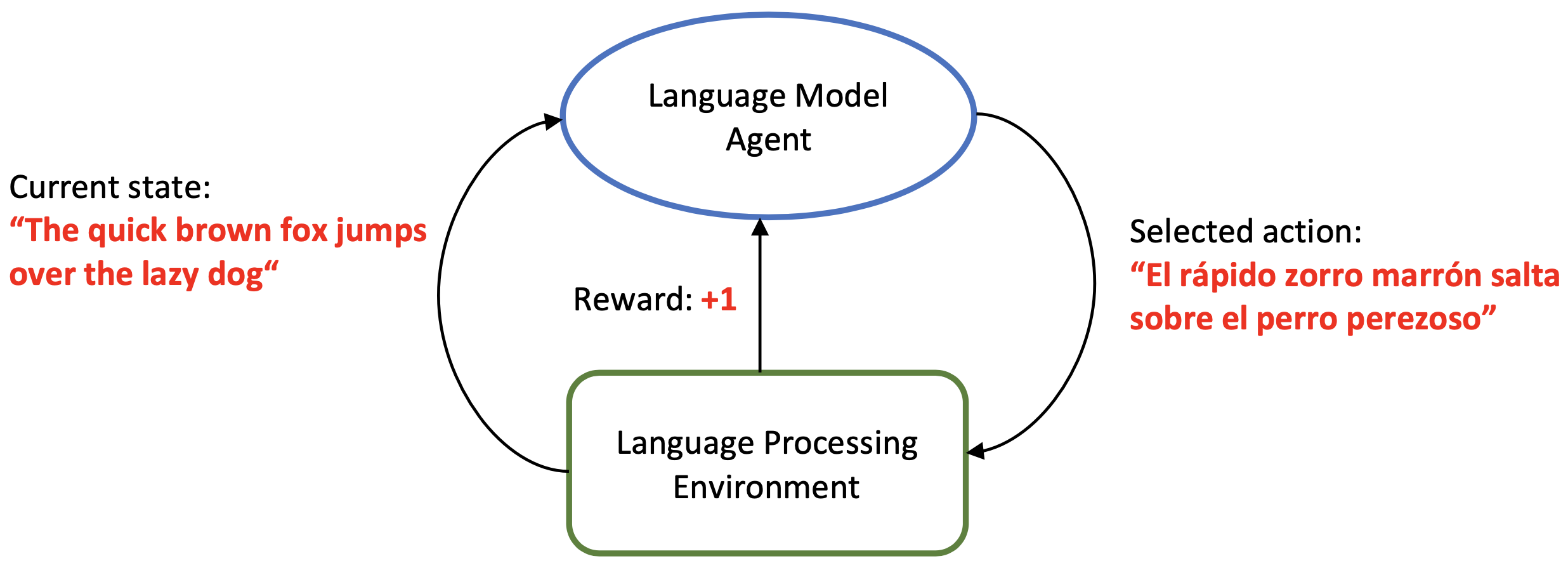}
 \caption{Schematic view of a reinforcement learning agent designed for language translation. It gets as input a text in some language A, and responds with another text string in a different language B. Input and output text strings have the same meaning. The language model agent acts by selecting the most relevant string of words. The language processing environment will provide the agent with the states and rewards after each of the interactions. The environment and the reward function are determined by the machine translation task being solved, i.e., translation from English to Spanish.}
\label{fig:translation}
 \end{figure}

Machine translation (MT) consists in automatically translating sentences from one natural language to another one, using a computing device \cite{Hutchins1992}. An MT system is a program that receives text (or speech) in some language as input and automatically generates text (or speech), with the same meaning, but in a different language (see Fig.\ \ref{fig:translation}). Early MT systems translate scientific and technical documents, while current developments involve online translation systems, teaching systems, among others. MT systems have been successfully applied to an increasing number of practical problems \cite{Way2018}. Since 1949, when the task of machine translation was proposed to be solved using computers~\cite{Weaver1949}, several approaches have been studied over the years.
 
Statistical machine translation (SMT) is by far the most studied approach to machine translation. In this paradigm, translations are generated using statistical models whose parameters are estimated through the analysis of many samples of existing human translations, known as bilingual text corpora \cite{Brown1990,Koehn2010,Williams2016}. SMT algorithms are characterized by their use of machine learning methods, where neural networks have been used with some success \cite{Cho2014,Devlin2014,Kalchbrenner2013}.

\begin{figure}[ht]
 \centering
 \includegraphics[width=11cm]{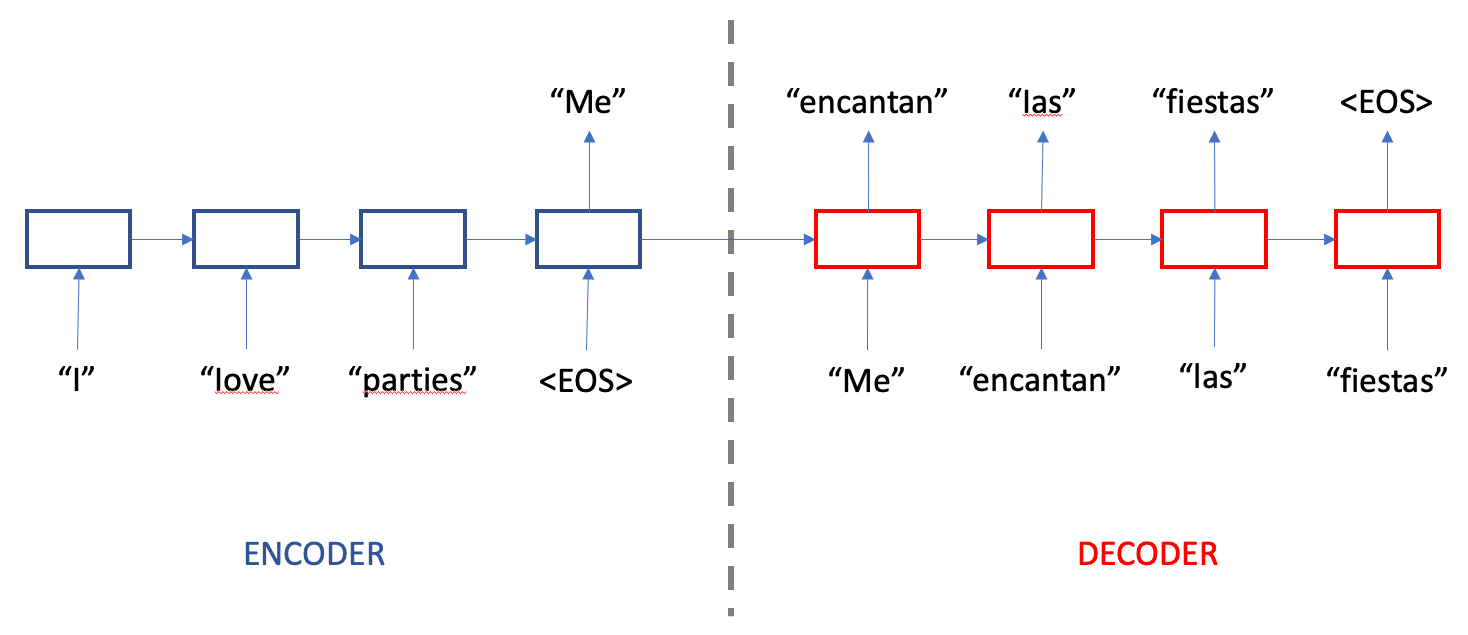}
 \caption{Sequence-to-sequence RNN architecture for machine translation, adapted from~\cite{Sutskever2014}.}
 \label{seq2seq}
 \end{figure}

In the last decade neural networks have won the battle against statistical methods in the field of translation. Neural Machine Translation (NMT) \cite{Stahlberg2020Neural} uses large neural networks to predict the likelihood of a sequence of words. NMT methods have been broadly applied to advance up-to-date phrase-based SMT systems, where a unit of translation may be a sequence of words (instead of a single word), called a phrase~\cite{Koehn2003}. NMT systems became a major area of development since the emergence of deep neural networks in 2012~\cite{Bahdanau2014,Wu2016,NIPS2017,Hassan2018,Lam2019}. Current state-of-the-art machine learning translation systems rely heavily on recurrent neural networks (RNN), such as the Long Short-Term Memory (LSTM) network ~\cite{Hochreiter1997}.
In the sequence-to-sequence approach~\cite{Sutskever2014} depicted in Fig.~\ref{seq2seq}, which was used for translation~\cite{Wu2016}, two recurrent neural networks are needed, an encoder and a decoder. The encoder RNN updates its weights as it receives a sequence of input words in order to extract the meaning of the sentence. Then, the decoder RNN updates its corresponding weights to generate the correct sequence of output words, in this case, the translated sentence. 
In the RNN approach the encoder makes reference to a program that would internally encode or represent the meaning of the source text, meanwhile the decoder will decode that internal representation and output a translated sentence with the correct meaning.
There are two problems that arise in the training and testing of seq2seq models. These problems are known as 1) exposure bias, i.e., the discrepancy between ground-truth dependent prediction during training and model-output dependent prediction during testing, and 2) inconsistency between the training and test objectives, i.e., measurement. Both problems have been recently studied and various solutions based on RL have been proposed \cite{Keneshloo2019}.

Similarly to what can be accomplished in conversational systems, in machine translation, we see that reinforcement learning algorithms can be used to predict the next word or phrase to be uttered by a person, specially during a simultaneous translation task, where the content is translated in real-time as it is produced~\cite{Fuegen2007}. This prediction is useful to increase the quality and speed up the translation.

In the case of the training, when it is done interactively, there is evidence that reinforcement learning can be used to improve the real-time translation performance after several interactions with humans \cite{grissomii2014,sokolov2015,sokolov2016}. Gu et al.\ \cite{Gu2017} propose an NMT model for real-time translation, where a task-specific neural network learns to decide which actions to take (i.e., to wait for another source word or to emit a target word) using a fixed pre-trained network and policy gradient techniques. Furthermore, to tackle the need of massive training data in machine translation, He et al.\ propose a dual learning mechanism, which automatically learns from unlabeled data \cite{He2016a}. This method is based on the fact that using a policy gradient algorithm together with a reward function defined as the likelihood of the language model, it is possible to create a translation model using examples of translation going in both directions, from language one to language two, and from language two to language one. With this approach it is possible to obtain an accuracy similar to the accuracy obtained with other neural models, but using only 10\% of the total number of training examples.

Speech translation systems have improved recently due to simultaneous machine translation, in which translation starts before the full sentence has been observed. In traditional speech translation systems, speech recognition results are first segmented into full sentences, then machine translation is performed sentence-by-sentence. However, as sentences can be long, i.e., in the case of lectures or presentations, this method can cause a significant delay between the speaker's utterance and the translation results, forcing listeners to wait a noticeable time until receiving the translation. Simultaneous machine translation avoids this problem by starting to translate before the sentence boundaries are detected. As a first step in this direction, Grissom II et al.\ \cite{grissomii2014} propose an approach that predicts next words and final verbs given a partial source language sentence by modeling simultaneous machine translation as a Markov decision process and using reinforcement learning. The policy introduced in this method works by keeping a partial translation, querying an underlying machine translation system and deciding to commit these intermediate translations occasionally. The policy is learned through the iterative imitation learning algorithm SEARN \cite{daume2009}. By letting the policy predict in advance the final verb of a source sentence, this method has the potential to notably decrease the delay in translation from languages in which, according to their grammar rules, the verb is usually placed in the end of the phrases, such as German. However, the successful use of RL is still very challenging, especially in real-world systems using deep neural networks and huge datasets \cite{Wu2018}.

Reinforcement learning techniques have also had a positive impact in statistical machine translation, which uses predictive algorithms to teach a computer how to translate text based on creating the most probable output learned from different bilingual text corpora. As the goal in reinforcement learning is to maximize the expected reward for choosing an action at a given state in an MDP model, algorithms based on bandit feedback for SMT can be visualized as MDP's with one state, where selecting an action represents the prediction of an output \cite{Langford2007}, \cite{Li2010}. Bandit feedback inherits the name from the problem of maximizing the amount of rewards obtained after a sequence of plays with a one-armed bandit machine, without apriori knowledge of the reward distribution function of the bandit machine. Sokolov et al.\ \cite{sokolov2015} propose a structured prediction in SMT based on bandit feedback, called \emph{bandit expected loss minimization}. This approach uses stochastic optimization for learning from partial feedback in the form of an expected 1--BLEU loss criterion \cite{Och2003}, \cite{Wuebker2015}, as opposed to learning from a gold standard reference translation. This is a non-convex optimization problem, which they analyzed in the stochastic gradient method of pseudogradient adaptation \cite{Poljak1973} that allowed to show convergence of the algorithm. Nevertheless, the algorithm of Sokolov et al.\ \cite{sokolov2015} presents slow convergence. In other words, such a system needs many rounds of user feedback in order to learn in a real-world SMT. Moreover, it requires absolute feedback of translation quality. Therefore, Sokolov et al.\ \cite{sokolov2016} propose improvements with a strong convexification of the learning objective, formalized as bandit cross-entropy minimization to overcome the convergence speed problem. They also propose a learning algorithm based on pairwise preference rankings, which simplifies the feedback information.

The same approach used for machine translation can be used in a rephrasing system \cite{guo2015}. This system receives a sentence as an input, creates an internal representation of the information contained in such a sentence and then generates a second sentence with the same meaning of the first one.  The algorithms used to solve such a challenging problem are the long short-term memory (LSTM) and a deep Q network (DQN). The former is used to learn the representation of the input sentence and the latter is used to generate the output sentence. The experiments presented in this work indicate that the proposed method performs very well at decoding sentences. Furthermore, the algorithm significantly outperformed the baseline when it was used to decode sentences never seen before, in terms of BLEU scores. The generation of the output string is not explicitly computed from a vector of variables, instead, this vector representation is implicitly learned and stored in the weights of the LSTM and the deep Q network. Similarly, this system does not need an explicit model of the language to do the rephrasing, because that model is also learned and stored in its neural networks. Therefore, the inner workings of this system are the same as a machine learning translator. It receives a string of words as input and generates another string of words with the same meaning.  

The rephrasing problem aforementioned consists in generating one string $B$ based on some input string $A$, in such a way that both strings have the same meaning. Considering this task we can define an MDP $(S,A,P,R)$ as proposed in \cite{guo2015}:

\begin{itemize}
\item The set of states $S$ is defined as the set of all possible input strings $w_i$. 
\item The set of actions $A$ consists of adding and deleting words taken from some vocabulary. 
\item The transition function $P$ can be completely determined and it is deterministic. The next state is the string that results from adding or deleting a word. 
\item Finally, the reward function $R$ can be defined as a function that measures how similar the strings $A$ and $B$ are, in semantical terms. 
\end{itemize}  

In general, machine translation can be defined as an optimization problem. In the particular case of simultaneous translation, we can define an MDP $(S,A,P,R)$ and solve it using reinforcement learning as we explain next. Given an utterance in a language $A$, we need to find the optimal utterance $B$ that maximizes a measure of semantic similarity with respect to $A$. In this kind of translation problem, when the sentences need to be translated as fast as possible, reinforcement learning can be used for learning when a part of a sentence should be trusted and used to translate future parts of the same sentence. In this way the person waiting for the translated sentence does not need to wait until the translator gets the last word of the original sentence to start the translation process. Therefore, the translation process can be accelerated by predicting the next noun or verb. The corresponding MDP is the following \cite{grissomii2014}:

\begin{itemize}
\item Each state in $S$ contains the string of words already seen by the translator and the next predicted word.
\item The actions in $A$ are mainly of three types: to commit to a partial translation, to predict the next word, or to wait for more words.
\item The transition function $P$, indicating the transitions from one state to another is fully determined by the current state and the action performed. We can compute the resulting string after applying an action.
\item The reward function $R$ can be defined based on the BLEU score~\cite{Papineni2002}, which basically measures how similar one translation is compared to a reference string, which is assumed to be available for training.
\end{itemize}

There is a number of improvements that could be researched in simultaneous machine translation using reinforcement learning. One is the implementation of these systems in more realistic scenarios where faster convergence is required. Currently, the experimentation with this approach has involved idealized situations in which the phrase to be translated contains only one verb. This constraint should be dropped if we want to employ them in real-world scenarios.
 
Experiments with other languages are also needed, especially for those languages that do not fall into the set of most spoken languages in the world. This will require the estimation of different optimal MDP policies, one for each language. However, if the correct recurrent neural model can be defined, using reinforcement learning might help in autonomously learning machine translation. In the same way that AlphaGo managed to play multiple games against itself and improved in the process, it might be the case that future translator algorithms can learn multiple natural languages by talking to themselves.

\section{Conversational systems}
\label{conversational}
Conversational systems are designed to interact with various users using natural language,  most commonly in verbal or written form. They are well structured and engineered to serve for instance as automated web assistance or for natural human-robot interaction. The architecture and functionality of such systems are heavily dependent on the application. 

There are two classes of conversational systems. First, open domain systems, usually known as chatbots. They are built in a Turing-test fashion. This is, they can hold a conversation basically about any topic, or at least they are trained with that goal in mind. Second, closed domain systems which are developed more as expert systems, in the sense that they should serve a conversational purpose very well defined and bounded. They should be able to provide information or assistance about a specific topic. In this article we are more interested in this latter system, since serving a well-defined task, can more easily benefit from reinforcement learning, due to reduced state and action spaces.

In this section, we will see that reinforcement learning algorithms can be used to generate suitable responses during a conversation with a human user. If the system can be programmed to predict with some accuracy how a conversation might occur, then it can optimize the whole process in such a way that the system can provide more information in less interactions if we are talking about a system designed to inform humans, or it can make a more interesting conversation if it is designed as a chatbot for entertainment. There are a number of factors that affect the effectiveness of a conversational system, including context identification, dynamic context adaptation, user intention \cite{crook2014}, and domain knowledge \cite{higashinaka2015}.
 
Conversational systems consist of three basic components whose sophistication will vary from system to system. These components are:

\begin{enumerate}
\item processing of the input message (perception), 
\item the internal state representation (semantic decoder), and 
\item the actions (dialogue manager).
\end{enumerate}

\begin{figure}[ht]
 \centering
 \includegraphics[width=11cm]{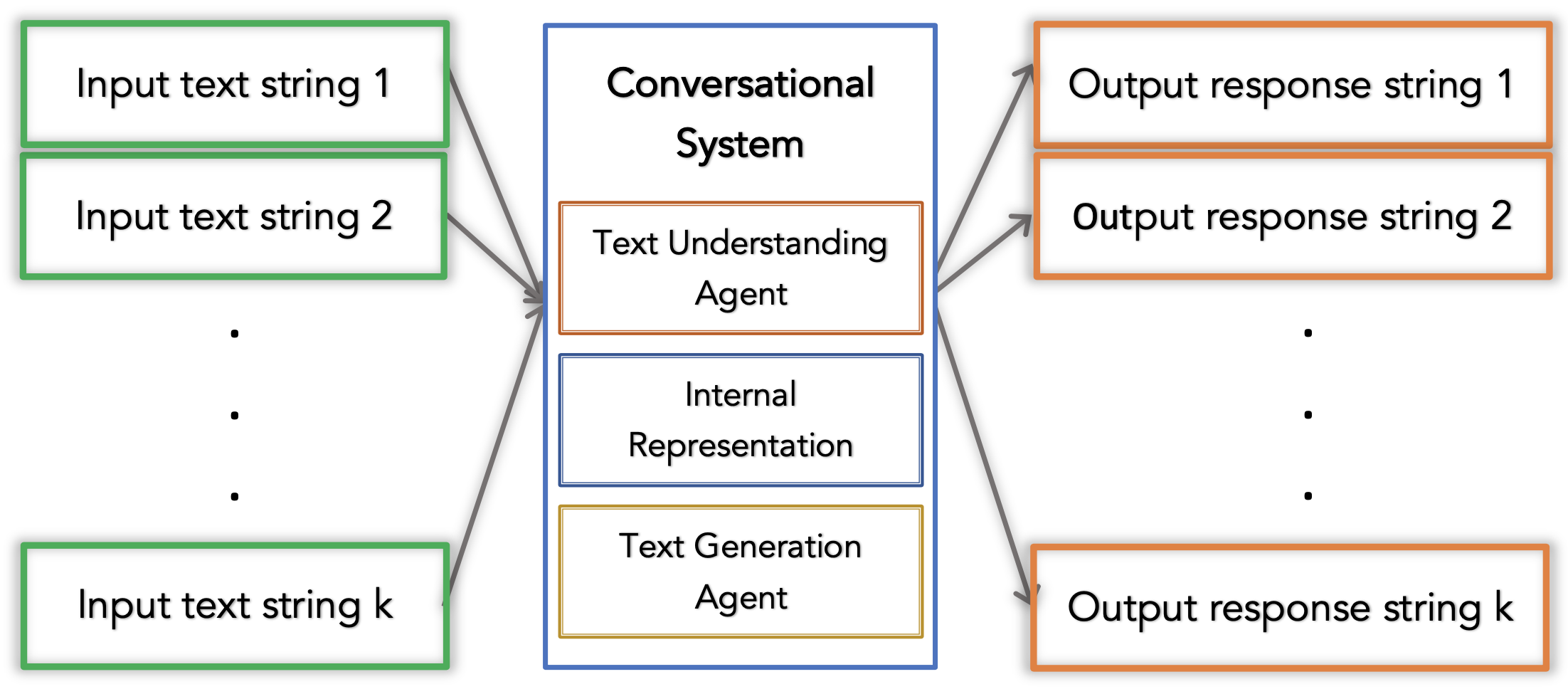}
 \caption{Information flow of a conversational system. This system receives as input a text string containing a question or simply a comment, and it responds with another text string containing the response. This input and response interaction typically iterates several times. Going from ``Input text string x'' to ``Output response string x'' requires the application in sequence of a text understanding agent (see Fig.\ \ref{parsing_scheme}) and a text generator agent (see Fig.\ \ref{text_generator_agent}).}
 \end{figure}

The input is a message from the user, for instance, speech, gestures, text, etc. The user's input message is converted to its semantic representation by the semantic encoder. The semantic representation of the message is further processed to determine an internal state of the system from which the next action is determined by the dialogue manager. Finally, the actions might include the generation of natural speech, text or other system actions. 

Conversational systems are often heuristically-driven and thus the flow of conversation as well as the capabilities are specifically tailored to a single application. Application-specific rule-based systems can achieve reasonably good performance due to the incorporation of expert domain knowledge. However, this often requires a huge number of rules, which becomes quickly intractable \cite{higashinaka2015}.

Due to the limitations of rule-based systems there are ongoing efforts to use data-driven or statistical conversational systems based on reinforcement learning since the early 2000s \cite{litman2000,levin2000,singh2000,singh2002,walker2000,young2000}. In theory, these data-driven conversational systems are capable of adapting based on interactions with real users. Additionally, they require less development effort but at a cost of significant learning time. Although very promising they still need to overcome several limitations before they are adopted for real-world applications. These limitations stem from both the problem itself and from reinforcement learning algorithms. 

Reinforcement learning could potentially be applied to all three components of a conversational system mentioned above, starting with perception of the input message, internal system representations as well as the decision of the system's output. However, we argue that reinforcement learning is more readily available for improving the dialogue manager which deals directly with the user interaction. More difficult but also possible using deep RL would be the learning of suitable internal representations based on the success of the interactions.

In a recent survey on neural approaches to conversational AI \cite{Gao2018}, it is recognized that in the last few years, reinforcement learning together with deep learning models have helped to significantly improve the quality of conversational agents in multiple tasks and domains. Key aspects of this combination of learning models are that conversational systems are allowed to adapt to different environments, tasks, domains and even user behaviors.

A large body of research exists for reinforcement learning-based conversational systems. For instance, POMDP-based conversational systems \cite{williams2007,young2010,thomson2010,young2013,crook2014} emerged as a strategy to cope with uncertainty originating from the perceptual and semantic decoder components. However, they also suffer from very large state representations that often become intractable (\textit{curse of dimensionality}) which typically necessitates some sort of state space compression \cite{crook2014}. We attribute this limitation to the widespread use of discrete state space representations typical in dialogue management and early days of reinforcement learning algorithms. We believe that such limitation could be overcome with continuous state space representations and the use of function approximation techniques such as DQN \cite{Mnih2015}, VIN \cite{tamar2016}, A3C \cite{mnih2016}, TRPO \cite{schulman2015} and many others. Although there have been attempts to use function approximation techniques within dialogue management systems \cite{jurcicek2010,henderson2008}, these have not been scaled up. Li et al.\ \cite{JiweiLi2016} simulated a dialogues between two virtual agents, and sequences that display three useful conversational properties are rewarded. These properties are: informativity, coherence, and ease of answering. This RL model uses policy gradient methods.

The main implications of using a continuous representation of the states is that we are required to estimate less parameters than when we use a discrete state representation. This is the case when we are dealing with large state spaces. As a result of handling less parameters the learning of policies can be significantly accelerated. Moreover, the quality of the learned policies is usually better than the policies learned with discretized state spaces. When we are implementing deep reinforcement learning models the number of weights in our neural network used to store the value functions can be large. However, the number of parameters of a deep model is less than the number of discrete states for which we would need to estimate a value.

Lemon et al.\ \cite{lemon2011} showed that natural language generation problems can be solved using reinforcement learning by jointly optimizing the generation of natural language and the management of dialogues. Another approach  based on RL to improve the long-turn coherence and consistency of a conversation is proposed in \cite{yu2017}. With this approach it is possible to obtain smooth transitions between task and non-task interactions. Papaioannou and Lemon \cite{papaioannou2017} present a chatbot system for task-specific applications. This system for multimodal human-robot interaction can generate longer conversations than a rule-based algorithm. This implies that the learned policy is highly successful in creating an engaging experience for chat and task interactions. A conversational agent can be effectively trained using a simulator \cite{liX2016}. After a preliminary training, the agent is deployed in the real scenario in order to generate interactions with humans. During these interactions with the real world the agent keeps learning. In a similar approach, Li et al.\ \cite{liX2017} used a movie booking system to test a neural conversational system trained to interact with users by providing information obtained from a structured database. Interestingly, if the action spaces of the agents are treated as latent variables, it is possible to induce those action spaces from the available data in an unsupervised learning manner. This approach can be used to train dialogue agents using reinforcement learning \cite{Zhao2019Rethinking}.

Some researchers have tried to develop question answering (QA) systems with multi-step reasoning capabilities, based on reinforcement learning. Though QA systems cannot be considered full conversational systems, both share some common challenges. DeepPath \cite{Xiong2017}, MINERVA \cite{Das2017} and M-Walk \cite{Shum2018} are recent examples of systems that perform multi-step reasoning on a knowledge base through the use of reinforcement learning.

More recently, Yang et al.\ \cite{Yang2020} presented a dialogue system that learns a policy that maximizes a joint reward function. The first reward term encourages topic coherence by computing the similarity between the topic representation of the generated response and that of the conversation history. The second term encourages semantic coherence between the generated response and previous utterance by computing mutual information. The last term is based on a language model to estimate the grammatical correctness and fluency of the generated response. Lu et al.\ \cite{Lu2019} used Hindsight Experience Replay (HER) to address the problem of sparse rewards in dialogues. HER allows for learning from failures and is thus effective for learning when successful dialogues are rare, particularly early in learning. Liu et al.\ \cite{Liu2020} showed that the goal is to model understanding between interlocutors rather than to simply focus on mimicking human-like responses. To achieve this goal, a transmitter-receiver-based framework is proposed. The transmitter generates utterances and the receiver measures the similarity between the built impression and the perceived persona. Mutual persona perception is then used as a reward to learn to generate personalized dialogues. Chen et al.\ \cite{ZChen2020} proposed a structured actor-critic model to implement structured deep reinforcement learning. It can learn in parallel from data taken from different conversational tasks,  achieving stable and sample-efficient learning. The method is tested on 18 tasks of PyDial \cite{Ultes2017pydial}. Plato et al.\ \cite{Plato2019,Plato2020} presented a complete attempt at concurrently training conversational agents. Such agents communicate only via self-generated language, outperforming supervised and deep learning baselines. Each agent has a role and a set of objectives, and they interact using only the language they have generated.

One major problem regarding the building of conversational systems lies in the amount of training data needed \cite{cuayahuitl2014} which could originate from simulations (as in most of the research), offline learning (limited number of interaction data sets) and learning from interactions with real users. In fact, training and evaluating such systems require large amounts of data. Similarly, measuring the performance of conversational systems is itself a challenge and different ways of measuring it have been proposed. One way is based on the use of some predefined metrics that can be used as the reward function of the system, for example, some measurement of the success rate of the system, which can be calculated when the system solves the user's problem. Another way of giving reward to the system is by counting the number of turns, which gives preference to more succinct dialogues. A more sophisticated way would be to automatically assess the sentiment of the evolving conversation, generating larger rewards for positive sentiment \cite{Bothe2017}. Other metrics that are being explored are the coherence, diversity and personal style of a more human-like conversational system \cite{Gao2018}.

Another way of measuring the performance is through the use of human simulators. However,  programming human simulators is not a trivial task. Moreover, once we have found a functional dialogue policy, there is no way to evaluate it without relying on heuristic methods. Some simulators are completely built from available data. The way they work is basically by selecting at the start of each training episode a randomly generated goal and a set of constraints. The performance of the system is measured by comparing the sequence of contexts and utterances generated after each step during the training. User simulation is not obvious and is still an ongoing research field. 

In general, conversational systems can be classified into two different types:
1) task-oriented systems, and 2) non-task-oriented systems.
Both types of systems can be defined as a general optimization problem that can be solved using reinforcement learning algorithms. An MDP $(S,A,T,R)$ with the main elements required to solve such an optimization problem is the following:

\begin{itemize}
\item The set of states $S$ is defined as the history of all utterances, such as comments, questions and answers happening during the dialogue. 
\item The set of actions $A$ consists of all the possible sentences that the system can answer to the user in the next time step.
\item The transition function $T$. The next state is the updated history of utterances after adding the last sentence generated by the system or the user. The transition function is non-deterministic in the case of non-predictable user responses.
\item Finally, the reward function $R$ can be defined as a function that measures the performance of the system, or how similar the generated dialogue is with respect to a reference dialogue from an existing corpus.
\end{itemize} 

The training of conversational systems could be also done using human users or using a model learned from corpora of a human-computer dialogue. However, the large number of possible dialogue states and strategies makes it difficult to be explored without employing a simulator. Therefore, the development of reliable user simulators is imperative for building conversational systems, and this comes with its own set of challenges.

Simulators are in particular useful for getting effective feedback from the environment during learning. For instance, Schatzmann et al.\ \cite{Schatzmann2009Hidden} implemented a user simulator using a stack structure to represent the states. The dialogue history in this approach consists of sequences of push and pop operations. Experiments show the effectiveness of this method to optimize a policy and it was shown to outperform a hand-crafted baseline strategy, in a real-world dialogue system. However, using a simulator always has serious limitations, whether it is manually coded, learned from available data, or a mixture of these approaches. A simulator is by definition not the real environment and therefore a reinforcement learning policy trained on it will need some or many adjustments to make it work properly in the real environment. In general, the development of realistic simulators for reinforcement learning and the related methodologies to fine-tune the policies afterwards to make them generalize well in the real world is still an open question. Moreover, the reward function is key to providing effective feedback. It is well known that the design of reward functions is a challenging task that requires expert knowledge on the task to be learned and on the specific algorithm being used. Very often, it is only after many iterations in the design process and a significant amount of experimentation that reward functions are optimally configured. Su et al.\ studied reward estimation \cite{Su2018Reward}. This approach is based on the one hand on the use of a recurrent neural network pre-trained off-line to serve as a predictor of success and on the other hand, a dialogue policy and a reward function are trained together. The reward function is modeled with a Gaussian process using active learning.

Chen et al.\ propose an interactive reinforcement learning framework to address the cold start problem \cite{Chen2017OnLine}. The framework, referred to as a companion teacher, consists of three parties: 1) one learning agent, 2) a human user, and 3) a human `companion' teacher. The agent (dialogue manager) consists of a dialogue state tracker and a policy model. The human teacher can guide learning at every turn (time step). The teacher can guide learning by both reward or policy-shaping. The authors assume that the dialogue states and policy model are visible to the human teacher. In follow-up work \cite{Chen2017AgentAware}, a rule-based system is used for reward- and policy-shaping, but the same strategy could be used to incorporate human feedback. The learning agent is implemented using a Deep Q-Network (DQN) and two separate experience memories for the agent and teacher. Uncertainty estimation is used to control when to ask for feedback and learn from the experience memories. Simulation experiments showed that the proposed approach could significantly improve learning speed and accuracy.

\section{Other language processing tasks}

Reinforcement learning has also been used for the improvement of information extraction through the acquisition and incorporation of external information \cite{Narasimhan2016Improving}. In this work, a deep Q-network is trained to select actions based on contextual information, leading the information retrieval system to improve its performance by increasing the accuracy of the retrieved documents. This approach can help to reduce the ambiguity in text interpretation. The selection of actions involves querying and extracting new sources of information repetitively. Actions have two components, a reconciliation decision and a query choice. The reward is designed to maximize the extraction accuracy of the values, and at the same time the number of queries is minimized. The experimental work with two domains shows an improvement over traditional information extractors of 5\% on average.

News feed recommendation can be seen as a combinatorial optimization problem and therefore it can be modeled as a Markov decision process. He et al.\ \cite{He2016Deep} studied the prediction of popular Reddit threads using a bi-directional LSTM architecture and reinforcement learning. Another approach to the same problem involves the incorporation of global context available in the form of discussions from an external source of knowledge \cite{He2017Reinforcement}. An interesting idea explored in this approach is the use of two Q-functions. The first is used to generate a first ranking of the actions and the second one is utilized to rerank top action candidates. By doing this, good actions can be selected, i.e., These actions could otherwise be missed due to the very skewed action space that the algorithm can deal with.

Quite often we see that dialogue systems provide semantically correct responses which are not necessarily consistent with contextual facts. Mesgar et al.\ \cite{mesgar-etal-2021-improving} used reinforcement learning to fine-tune the responses, optimizing for consistency and semantics.

Gao et al.\ \cite{Gao2019RewardLearning} approached another language processing task using reinforcement learning, namely document summarization. The proposed paradigm uses learning-to-rank as a way to learn a reward function that is later used to generate near-optimal summaries.

\section{Promising research directions}
\label{directions}
Based on our analysis of the problems and approaches here reported, we now take a step further and describe 9 research directions that we believe will benefit from a reinforcement learning approach in the coming years.

\begin{enumerate}

\item {\bf Recognition of the user's input.}
We noticed that a common element missing or at least underrepresented in natural language processing research is the recognition of the user's input. Commonly, this is treated as being inherently uncertain and most research accepts this and tries to cope with it without attempting to solve the source of the problems. This along with all other machine perception problems are very challenging tasks and far from being solved. We argue that trying to address uncertainty of the user input at the initial stages would be more fruitful than simply regarding it as given. Thus, we argue that a future research direction would be to develop a reinforcement learning approach for generating internal semantic representations of the user's message from which other fields within and beyond natural language processing could benefit. 

\item {\bf Internal representation learning.}
Learning an internal representation of language is a more general research direction. By using deep neural networks and reinforcement learning methods, it is possible to learn to code and decode sequences of text \cite{guo2015}. Although such an architecture was implemented and tested only with a text rephrasing task, we believe that the underlying problem of learning an internal representation of language is inherently related to some of the most important NLP problems, such as text understanding, machine translation, language generation, dialogue system management, parsing, etc. By solving the internal representation problem of language, we may partially solve the aforementioned problems to some extent. Therefore, research on deep learning and reinforcement learning methods in a joint approach is currently of great importance to advance the state of the art in NLP systems.

\item {\bf Exploitation of domain knowledge.}
Another interesting research path is the one aiming at discovering ways to enhance RL through the exploitation of domain knowledge available in the form of natural language, as surveyed by Luketina et al.\ \cite{Luketina2019ASO}. Some current trends involve methods studying knowledge transfer from descriptive task-dependent language corpora \cite{Narasimhan2018}. Pre-trained information retrieval systems can be integrated with RL agents \cite{chen2017} to improve the quality of the queries. Moreover, relevant information can be extracted from sources of unstructured data such as game manuals \cite{branavan2012}.

\item {\bf Exploitation of embodiment.}
A trend in supervised language learning research considers the importance of embodiment for the emergence of language \cite{antunes2019mtlstm,heinrich2020}.
Multimodal inputs, such as an agent knowing its actuators while performing an action, help in classifying and verbally describing an action and allows better generalisation to novel action-object combinations \cite{eisermann2021}.
Embodied language learning has recently been brought to reinforcement learning scenarios, specifically question answering where an agent needs to navigate in a scene to answer the questions \cite{tan2020embodied}, or where it needs to perform actions on objects to answer questions \cite{deng2020mqa}.
Like dialogue grounded in vision \cite{das2017learning}, such interactive scenarios extend language learning into multiple modalities. Such applied scenarios also allow to introduce tasks, corresponding rewards, and hence seamless integration of language learning with reinforcement learning.
Deep reinforcement learning neural architectures are a promising research path for the processing of multiple modalities in embodied language learning in a dynamic world.

\item {\bf Language evolution.}
From a more linguistic point of view, the study of language evolution using a reinforcement learning perspective is also a fertile field for research. This process can be modelled by a multi-agent system, where a collection of agents is capable to create their own communication protocol by means of interaction with a common environment and by applying reinforcement learning rules \cite{Mordatch2018}. This kind of research can benefit from the recent advances in multi-agent systems and rising computational power. Moreover, research on cognitive robotics using neural models together with reinforcement learning methods \cite{CPW18,CMNW18,RENW20,ENW19,HWKW19} has reached a point where the addition of language evolution capabilities seems to be more promising than ever before.

\item {\bf Word embeddings.}
More important, from our point of view, are the advances in neural language models, especially those for word embedding. The recent trend of continuous language representations might have a huge potential if it is used together with reinforcement learning. Word2vec \cite{Mikolov2013} supplies a continuous vector  representation of words. In a continuous bag-of-words architecture,  Word2vec trains a simple neural network to predict a word from its surrounding words, achieving on its hidden layer a low-dimensional  continuous representation of words in some semantically meaningful  topology. Other word embeddings are GloVe \cite{Pennington2014}, which  yields a similar performance more efficiently by using a co-occurrence  matrix of words in their context, and FastText \cite{Bojanowski2017},  which includes subword information to enrich word vectors and to deal  with out-of-vocabulary words.

A more powerful class of embeddings are contextualized word  embeddings, which use the context, i.e., previous and following words,  to embed a word. Two recent models are ELMo \cite{Peters2018}, which uses bidirectional LSTM, and BERT \cite{Devlin2018}, which uses a deep  feedforward Transformer network architecture with self-attention. Both  are character-based and hence, like FastText, use morphological cues  and deal with out-of-vocabulary words. By taking into account the context, they handle different meanings of a word (e.g., 
``He touches a  rock'' vs ``He likes rock''). However, simple word embeddings become  meaning embeddings, blurring the distinction between word- and  sentence embeddings.

For the representations of utterances, Word2vec has been extended to  Doc2vec \cite{LeMikolov2014}, and other simple schemes are based on a  weighted combination of contained word vectors  \cite{Arora2016,Ruckle2018}. However, since these simple  bag-of-words approaches lose word-order information, the original sentence cannot be reconstructed. Sentence generation is also difficult for supervised  sentence embeddings such as InferSent \cite{Conneau2017} or Google's  Universal Sentence Encoder \cite{Cer2018}.

An unsupervised approach to sentence vectors are Skip-Thought Vectors  \cite{Kiros2015}, which are trained to reconstruct the surrounding  sentences of an encoded one. A simpler model would be an  encoder-decoder autoencoder architecture, where the decoder  reconstructs the same utterance that the encoder gets as input, based  on a constant-length internal representation. Hence, this is a  constant size continuous vector representation of an utterance, from  which the utterance, which itself could consist of continuous word  vectors, could also be reproduced.

To train utterance vectors on dialogues, large dialogue corpora exist,  which can be classified into human-machine or human-human;  spontaneously spoken, scripted spoken, or written \cite{Serban2015}.  Examples are datasets of annotated telephone dialogues, movie  dialogues, movie recommendation dialogues, negotiation dialogues,  human-robot interaction, and also question answering contains elements  of dialogues.

Such continuous language representations could seamlessly play  together with continuous RL algorithms like CACLA \cite{VanHasselt2007},  Deterministic Policy Gradient (DPG) \cite{Silver2014} or deep DPG  (DDPG) \cite{Lillicrap2015}. These algorithms handle continuous state  input and continuous action output. Actions of a dialogue agent would  be the agent's utterances, which would result in a new state after the  response of its communication partner. Continuous utterance  representations would allow optimization of an action by gradient ascent to maximize certain rewards which express desired future state properties. For example, it could be desired to maximize the positive  sentiment of an upcoming utterance which can be estimated by a  differentiable neural network \cite{Bothe2017}.

Other possible desired state properties could be to maximize a human's  excitement in order to motivate him to make a decision; to maximize  the duration of the conversation, or lead it to an early end with a  pleased human; to acquire certain information from, or to pass on  information to the human. However, not all goals can be easily  expressed as points in a continuous utterance space that represents a  dialogue. To this end, future research on language needs to be  extended towards representing more of its semantics, which entails  understanding the entire situation.

\item {\bf Intelligent conversational systems.}
When conversing with chatbots, it is common to end up in the situation where the bot starts responding with “I don’t know what you are talking about” repeatedly, no matter what it is asked. This problem is identified as the generic response problem. The cause for this problem might be that such kind of answers occur very often in the training set. Also they are highly compatible with various questions \cite{JiweiLi2016}. Another issue is when a dataset has similar responses to different contexts \cite{Sankar2019}. One way to improve the efficiency in reinforcement learning is through the combination of model-based and model-free learning \cite{Hafez2020}. We propose that this approach might be useful to solve the generic response problem.

Furthermore, all the experience gained from working with algorithms designed for text-based games and applications on learning of navigational directions can be extended and adapted to be useful in the implementation of intelligent tutors, smart enough to understand the questions posed by the user and select the most appropriate learning resource, whether it is some text, audio, video, hyperlink, etc. Those intelligent tutors can improve over time.

\item {\bf Assessment of conversational systems.}
Finally, in conversational  systems, a critical point that needs further investigation is the definition of robust evaluation schemes that can be automated and used to assess the quality of automatic dialogue systems. Currently, the performance of such systems is measured through ad hoc procedures that depend on the specific application and most importantly, they require the intervention of a human, which makes these systems very difficult to be scaled.

\item {\bf Document-editing RL Assistants.}
Kudashkina et al.\ \cite{Kudashkina2020} proposed the domain of voice document editing as a particularly well-suited one for the development of reinforcement learning intelligent assistants that can engage in a conversation. They argue that in voice document editing, the domain is clearly defined, delimited and the agent has full access to it. These conditions are advantageous for an agent that learns the domain of discourse through model-based reinforcement learning. Important future research questions the authors mention are, first, what level of ambition should the agent's learning have? And second, how should the training of the assistant be performed, online or offline?

\end{enumerate}

\section{Conclusions}
\label{conclusions}
We have provided a review of the main categories of natural language processing problems that have been approached using reinforcement learning methods. Some of these problems considered reinforcement learning as the main algorithm, such as the dialogue management
systems. In others, reinforcement learning was used marginally, only to partially help in the solution of the central problem. In both cases, RL algorithms have played an important part in the optimization of control policies through the self-exploration of the states and actions.

With the current advances in reinforcement learning algorithms, especially with those algorithms in which the value functions and policy functions are replaced with deep neural networks, it is impossible not to consider that reinforcement learning will play a major role in solving some of the most important natural language processing problems. Especially, we have witnessed solid evidence that algorithms with self-improvement and self-adaptation capabilities have pushed the performance in challenging machine learning problems to the next level.

Currently, none of the natural language processing tasks here analyzed have reinforcement learning methods as state-of-the-art methodologies. Many of the problems are being solved with increasing success using transformer neural network models such as BERT and GPT. However, we argue that reinforcement learning can be jointly applied with deep neural models. Reinforcement learning can  provide benefit by its inherent exploratory capacity. This is, reinforcement learning can help find better actions and better states due to its credit assignment  approach. The best policies found by neural networks, such as transformers, can potentially get fine-tuned by reinforcements.

\section*{Acknowledgements}
This work received partial support from the German Research Foundation (DFG) under project CML (TRR-169). We thank Burhan Hafez for discussions and providing references highly relevant to this review.

\bibliographystyle{abbrv}       
\bibliography{rl_for_nlp}   

\begin{thebibliography}{100}

\bibitem{antunes2019mtlstm}
A.~Antunes, A.~Laflaquiere, T.~Ogata, and A.~Cangelosi.
\newblock A {{Bi}}-directional {{Multiple Timescales LSTM Model}} for
  {{Grounding}} of {{Actions}} and {{Verbs}}.
\newblock In {\em {{IEEE}}/{{RSJ International Conference}} on {{Intelligent
  Robots}} and {{Systems}} ({{IROS}})}, pages 2614--2621, {Macau, China}, Nov.
  2019.

\bibitem{Arora2016}
S.~Arora, Y.~Liang, and T.~Ma.
\newblock A {{Simple}} but {{Tough}}-to-{{Beat Baseline}} for {{Sentence
  Embeddings}}.
\newblock In {\em International {{Conference}} on {{Learning Representations}}
  ({{ICLR}})}, {Toulon, France}, Apr. 2017. {OpenReview.net}.

\bibitem{Bahdanau2014}
D.~Bahdanau, K.~Cho, and Y.~Bengio.
\newblock Neural {{Machine Translation}} by {{Jointly Learning}} to {{Align}}
  and {{Translate}}.
\newblock In {\em International {{Conference}} on {{Learning Representations}}
  ({{ICLR}})}, {San Diego, CA, USA}, May 2015. {arxiv}.

\bibitem{Bengio2015}
S.~Bengio, O.~Vinyals, N.~Jaitly, and N.~Shazeer.
\newblock Scheduled {{Sampling}} for {{Sequence Prediction}} with {{Recurrent
  Neural Networks}}.
\newblock In {\em International {{Conference}} on {{Neural Information
  Processing Systems}} ({{NIPS}})}, volume~1, pages 1171--1179, {Montreal, QC,
  Canada}, Dec. 2015. {MIT Press}.

\bibitem{Bojanowski2017}
P.~Bojanowski, E.~Grave, A.~Joulin, and T.~Mikolov.
\newblock Enriching {{Word Vectors}} with {{Subword Information}}.
\newblock {\em Transactions of the Association for Computational Linguistics},
  5:135--146, Dec. 2017.

\bibitem{Bothe2017}
C.~Bothe, S.~Magg, C.~Weber, and S.~Wermter.
\newblock Dialogue-{{Based Neural Learning}} to {{Estimate}} the {{Sentiment}}
  of a {{Next Upcoming Utterance}}.
\newblock In A.~Lintas, S.~Rovetta, P.~F. Verschure, and A.~E. Villa, editors,
  {\em International {{Conference}} on {{Artificial Neural Networks}}
  ({{ICANN}})}, volume 10614 of {\em Lecture {{Notes}} in {{Computer
  Science}}}, pages 477--485, {Alghero, Italy}, Sept. 2017. {Springer
  International Publishing}.

\bibitem{branavan2012}
S.~R.~K. Branavan, D.~Silver, and R.~Barzilay.
\newblock Learning to {{Win}} by {{Reading Manuals}} in a {{Monte}}-{{Carlo
  Framework}}.
\newblock {\em Journal of Artificial Intelligence Research}, 43:661--704, Apr.
  2012.

\bibitem{Brown1990}
P.~F. Brown, J.~Cocke, S.~A.~D. Pietra, V.~J.~D. Pietra, F.~Jelinek, J.~D.
  Lafferty, R.~L. Mercer, and P.~S. Roossin.
\newblock A {{Statistical Approach}} to {{Machine Translation}}.
\newblock {\em Computational Linguistics}, 16(2):79--85, June 1990.

\bibitem{brown2020language}
T.~B. Brown, B.~Mann, N.~Ryder, M.~Subbiah, J.~Kaplan, P.~Dhariwal,
  A.~Neelakantan, P.~Shyam, G.~Sastry, A.~Askell, S.~Agarwal,
  A.~{Herbert-Voss}, G.~Krueger, T.~Henighan, R.~Child, A.~Ramesh, D.~M.
  Ziegler, J.~Wu, C.~Winter, C.~Hesse, M.~Chen, E.~Sigler, M.~Litwin, S.~Gray,
  B.~Chess, J.~Clark, C.~Berner, S.~McCandlish, A.~Radford, I.~Sutskever, and
  D.~Amodei.
\newblock Language {{Models Are Few}}-{{Shot Learners}}.
\newblock In {\em Neural {{Information Processing Systems (NeurIPS)}}}, {Online
  Conference}, Dec. 2020.

\bibitem{Cangelosi2002}
A.~Cangelosi and D.~Parisi, editors.
\newblock {\em Simulating the {{Evolution}} of {{Language}}}.
\newblock {Springer-Verlag}, {London}, 2002.

\bibitem{CaoZhu2019}
R.~Cao, S.~Zhu, C.~Liu, J.~Li, and K.~Yu.
\newblock Semantic {{Parsing}} with {{Dual Learning}}.
\newblock In {\em Annual {{Meeting}} of the {{Association}} for {{Computational
  Linguistics}} ({{ACL}})}, volume 57th, pages 51--64, {Florence, Italy}, July
  2019. {Association for Computational Linguistics}.

\bibitem{Cer2018}
D.~Cer, Y.~Yang, S.-y. Kong, N.~Hua, N.~Limtiaco, R.~S. John, N.~Constant,
  M.~{Guajardo-Cespedes}, S.~Yuan, C.~Tar, Y.-H. Sung, B.~Strope, and
  R.~Kurzweil.
\newblock Universal {{Sentence Encoder}}.
\newblock {\em arXiv:1803.11175 [cs]}, Apr. 2018.

\bibitem{Che2017}
T.~Che, Y.~Li, R.~Zhang, R.~D. Hjelm, W.~Li, Y.~Song, and Y.~Bengio.
\newblock Maximum-{{Likelihood Augmented Discrete Generative Adversarial
  Networks}}.
\newblock {\em arXiv:1702.07983 [cs]}, Feb. 2017.

\bibitem{chen2017}
D.~Chen, A.~Fisch, J.~Weston, and A.~Bordes.
\newblock Reading {{Wikipedia}} to {{Answer Open}}-{{Domain Questions}}.
\newblock In {\em Annual {{Meeting}} of the {{Association}} for {{Computational
  Linguistics}} ({{ACL}})}, volume 55th, pages 1870--1879, {Vancouver, BC,
  Canada}, July 2017. {Association for Computational Linguistics}.

\bibitem{Chen2017OnLine}
L.~Chen, R.~Yang, C.~Chang, Z.~Ye, X.~Zhou, and K.~Yu.
\newblock On-{{Line Dialogue Policy Learning}} with {{Companion Teaching}}.
\newblock In {\em Conference of the {{European Chapter}} of the {{Association}}
  for {{Computational Linguistics}} ({{EACL}})}, volume 15th of {\em Short
  {{Papers}}}, pages 198--204, {Valencia, Spain}, Apr. 2017. {Association for
  Computational Linguistics}.

\bibitem{Chen2017AgentAware}
L.~Chen, X.~Zhou, C.~Chang, R.~Yang, and K.~Yu.
\newblock Agent-{{Aware Dropout DQN}} for {{Safe}} and {{Efficient}} on-{{Line
  Dialogue Policy Learning}}.
\newblock In {\em Conference on {{Empirical Methods}} in {{Natural Language
  Processing}} ({{EMNLP}})}, pages 2454--2464, {Copenhagen, Denmark}, Sept.
  2017. {Association for Computational Linguistics}.

\bibitem{ZChen2020}
Z.~Chen, L.~Chen, X.~Liu, and K.~Yu.
\newblock Distributed {{Structured Actor}}-{{Critic Reinforcement Learning}}
  for {{Universal Dialogue Management}}.
\newblock {\em IEEE/ACM Transactions on Audio, Speech, and Language
  Processing}, 28:2400--2411, 2020.

\bibitem{Cho2014}
K.~Cho, B.~{van Merri{\"e}nboer}, C.~Gulcehre, D.~Bahdanau, F.~Bougares,
  H.~Schwenk, and Y.~Bengio.
\newblock Learning {{Phrase Representations Using RNN
  Encoder}}\textendash{{Decoder}} for {{Statistical Machine Translation}}.
\newblock In {\em Conference on {{Empirical Methods}} in {{Natural Language
  Processing}} ({{EMNLP}})}, pages 1724--1734, {Doha, Qatar}, Oct. 2014.
  {Association for Computational Linguistics}.

\bibitem{chomsky1959certain}
N.~Chomsky.
\newblock On {{Certain Formal Properties}} of {{Grammars}}.
\newblock {\em Information and Control}, 2(2):137--167, June 1959.

\bibitem{chomsky65UG}
N.~Chomsky.
\newblock {\em Aspects of the {{Theory}} of {{Syntax}}}.
\newblock {The MIT Press}, {Cambridge, Mass}, May 1965.

\bibitem{Conneau2017}
A.~Conneau, D.~Kiela, H.~Schwenk, L.~Barrault, and A.~Bordes.
\newblock Supervised {{Learning}} of {{Universal Sentence Representations}}
  from {{Natural Language Inference Data}}.
\newblock In {\em Conference on {{Empirical Methods}} in {{Natural Language
  Processing}} ({{EMNLP}})}, pages 670--680, {Copenhagen, Denmark}, Sept. 2017.
  {Association for Computational Linguistics}.

\bibitem{crook2014}
P.~A. Crook, S.~Keizer, Z.~Wang, W.~Tang, and O.~Lemon.
\newblock Real {{User Evaluation}} of a {{POMDP Spoken Dialogue System Using
  Automatic Belief Compression}}.
\newblock {\em Computer Speech \& Language}, 28(4):873--887, July 2014.

\bibitem{CMNW18}
F.~Cruz, S.~Magg, Y.~Nagai, and S.~Wermter.
\newblock Improving {{Interactive Reinforcement Learning}}: {{What Makes}} a
  {{Good Teacher}}?
\newblock {\em Connection Science}, 30(3):306--325, Mar. 2018.

\bibitem{CPW18}
F.~Cruz, G.~I. Parisi, and S.~Wermter.
\newblock Multi-modal {{Feedback}} for {{Affordance}}-driven {{Interactive
  Reinforcement Learning}}.
\newblock In {\em International {{Joint Conference}} on {{Neural Networks}}
  ({{IJCNN}})}, pages 1--8, {Rio de Janeiro, Brazil}, July 2018.

\bibitem{cuayahuitl2014}
H.~Cuay{\'a}huitl, I.~{Kruijff-Korbayov{\'a}}, and N.~Dethlefs.
\newblock Nonstrict {{Hierarchical Reinforcement Learning}} for {{Interactive
  Systems}} and {{Robots}}.
\newblock {\em ACM Transactions on Interactive Intelligent Systems},
  4(3):15:1--15:30, Oct. 2014.

\bibitem{das2017learning}
A.~Das, S.~Kottur, J.~M.~F. Moura, S.~Lee, and D.~Batra.
\newblock Learning {{Cooperative Visual Dialog Agents}} with {{Deep
  Reinforcement Learning}}.
\newblock In {\em {{IEEE International Conference}} on {{Computer Vision}}
  ({{ICCV}})}, pages 2951--2960, {Venice, Italy}, Oct. 2017.

\bibitem{Das2017}
R.~Das, S.~Dhuliawala, M.~Zaheer, L.~Vilnis, I.~Durugkar, A.~Krishnamurthy,
  A.~Smola, and A.~McCallum.
\newblock Go for a {{Walk}} and {{Arrive}} at the {{Answer}}: {{Reasoning Over
  Paths}} in {{Knowledge Bases Using Reinforcement Learning}}.
\newblock In {\em International {{Conference}} on {{Learning Representations}}
  ({{ICLR}})}, {Vancouver, BC, Canada}, 2018.

\bibitem{daume2009}
H.~Daum{\'e}~III, J.~Langford, and D.~Marcu.
\newblock Search-{{Based Structured Prediction}}.
\newblock {\em Machine Learning}, 75(3):297--325, June 2009.

\bibitem{deng2020mqa}
Y.~Deng, X.~Guo, N.~Zhang, D.~Guo, H.~Liu, and F.~Sun.
\newblock {{MQA}}: {{Answering}} the {{Question}} via {{Robotic Manipulation}}.
\newblock {\em arXiv:2003.04641 [cs]}, Dec. 2020.

\bibitem{dethlefs2011a}
N.~Dethlefs and H.~Cuay{\'a}huitl.
\newblock Combining {{Hierarchical Reinforcement Learning}} and {{Bayesian
  Networks}} for {{Natural Language Generation}} in {{Situated Dialogue}}.
\newblock In {\em European {{Workshop}} on {{Natural Language Generation}}
  ({{ENLG}})}, volume~11, pages 110--120, {Nancy, France}, Sept. 2011.
  {Association for Computational Linguistics}.

\bibitem{dethlefs2011}
N.~Dethlefs and H.~Cuay{\'a}huitl.
\newblock Hierarchical {{Reinforcement Learning}} and {{Hidden Markov Models}}
  for {{Task}}-{{Oriented Natural Language Generation}}.
\newblock In {\em Annual {{Meeting}} of the {{Association}} for {{Computational
  Linguistics}}: {{Human Language Technologies}} ({{ACL}})}, volume~49 of {\em
  Short {{Papers}}}, pages 654--659, {Portland, OR, USA}, June 2011.
  {Association for Computational Linguistics}.

\bibitem{Devlin2018}
J.~Devlin, M.-W. Chang, K.~Lee, and K.~Toutanova.
\newblock {{BERT}}: {{Pre}}-{{Training}} of {{Deep Bidirectional Transformers}}
  for {{Language Understanding}}.
\newblock In {\em Conference of the {{North American Chapter}} of the
  {{Association}} for {{Computational Linguistics}}: {{Human Language
  Technologies}} ({{NAACL HLT}})}, pages 4171--4186, {Minneapolis, MN, USA},
  June 2019. {Association for Computational Linguistics}.

\bibitem{Devlin2014}
J.~Devlin, R.~Zbib, Z.~Huang, T.~Lamar, R.~Schwartz, and J.~Makhoul.
\newblock Fast and {{Robust Neural Network Joint Models}} for {{Statistical
  Machine Translation}}.
\newblock In {\em Annual {{Meeting}} of the {{Association}} for {{Computational
  Linguistics}} ({{ACL}})}, volume 52nd, pages 1370--1380, {Baltimore, MD,
  USA}, June 2014. {Association for Computational Linguistics}.

\bibitem{eisermann2021}
A.~Eisermann, J.~H. Lee, C.~Weber, and S.~Wermter.
\newblock Generalization in {{Multimodal Language Learning}} from
  {{Simulation}}.
\newblock In {\em International {{Joint Conference}} on {{Neural Networks}}
  ({{IJCNN}})}, pages 1--8, {Shenzhen, China}, 2021.

\bibitem{ENW19}
M.~Eppe, P.~D.~H. Nguyen, and S.~Wermter.
\newblock From {{Semantics}} to {{Execution}}: {{Integrating Action Planning
  With Reinforcement Learning}} for {{Robotic Causal Problem}}-{{Solving}}.
\newblock {\em Frontiers in Robotics and AI}, 6(123), Nov. 2019.

\bibitem{Fuegen2007}
C.~F{\"u}gen, A.~Waibel, and M.~Kolss.
\newblock Simultaneous {{Translation}} of {{Lectures}} and {{Speeches}}.
\newblock {\em Machine Translation}, 21(4):209--252, Dec. 2007.

\bibitem{Gao2018}
J.~Gao, M.~Galley, and L.~Li.
\newblock Neural {{Approaches}} to {{Conversational AI}}.
\newblock In {\em International {{ACM SIGIR Conference}} on {{Research}} \&
  {{Development}} in {{Information Retrieval}}}, volume 41st, pages 1371--1374,
  {Ann Arbor, MI, USA}, June 2018. {Association for Computing Machinery}.

\bibitem{Gao2019RewardLearning}
Y.~Gao, C.~Meyer, M.~Mesgar, and I.~Gurevych.
\newblock Reward {{Learning}} for {{Efficient Reinforcement Learning}} in
  {{Extractive Document Summarisation}}.
\newblock In {\em International {{Joint Conference}} on {{Artificial
  Intelligence}} ({{IJCAI}})}, 19th, pages 2350--2356, {Macao, China}, 2019.
  {AAAI Press}.

\bibitem{Goodfellow2014}
I.~Goodfellow, J.~{Pouget-Abadie}, M.~Mirza, B.~Xu, D.~{Warde-Farley},
  S.~Ozair, A.~Courville, and Y.~Bengio.
\newblock Generative {{Adversarial Nets}}.
\newblock In {\em Advances in {{Neural Information Processing Systems}}
  ({{NIPS}})}, volume~27, pages 2672--2680, {Montreal, QC, Canada}, Dec. 2014.
  {Curran Associates, Inc.}

\bibitem{grissomii2014}
A.~Grissom~II, H.~He, J.~{Boyd-Graber}, J.~Morgan, and H.~Daum{\'e}~III.
\newblock Don't {{Until}} the {{Final Verb Wait}}: {{Reinforcement Learning}}
  for {{Simultaneous Machine Translation}}.
\newblock In {\em Conference on {{Empirical Methods}} in {{Natural Language
  Processing}} ({{EMNLP}})}, pages 1342--1352, {Doha, Qatar}, Oct. 2014.
  {Association for Computational Linguistics}.

\bibitem{Gu2017}
J.~Gu, G.~Neubig, K.~Cho, and V.~O. Li.
\newblock Learning to {{Translate}} in {{Real}}-{{Time}} with {{Neural Machine
  Translation}}.
\newblock In {\em Conference of the {{European Chapter}} of the {{Association}}
  for {{Computational Linguistics}} (EACL)}, volume 15th, pages 1053--1062,
  {Valencia, Spain}, Apr. 2017. {Association for Computational Linguistics}.

\bibitem{guo2015}
H.~Guo.
\newblock Generating {{Text}} with {{Deep Reinforcement Learning}}.
\newblock In {\em {{NIPS}} Deep Reinforcement Learning Workshop}, {Montreal,
  QC, Canada}, 2015.

\bibitem{Guo2018}
J.~Guo, S.~Lu, H.~Cai, W.~Zhang, Y.~Yu, and J.~Wang.
\newblock Long {{Text Generation Via Adversarial Training}} with {{Leaked
  Information}}.
\newblock {\em Proceedings of the AAAI Conference on Artificial Intelligence},
  32(1):5141--5148, Apr. 2018.

\bibitem{guo2017}
X.~Guo, T.~Klinger, C.~Rosenbaum, J.~P. Bigus, M.~Campbell, B.~Kawas,
  K.~Talamadupula, G.~Tesauro, and S.~Singh.
\newblock Learning to {{Query}}, {{Reason}}, and {{Answer Questions}} on
  {{Ambiguous Texts}}.
\newblock In {\em International {{Conference}} on {{Learning Representations}}
  ({{ICLR}})}, {Toulon, France}, Apr. 2017.

\bibitem{HWKW19}
M.~B. Hafez, C.~Weber, M.~Kerzel, and S.~Wermter.
\newblock Deep {{Intrinsically Motivated Continuous Actor}}-{{Critic}} for
  {{Efficient Robotic Visuomotor Skill Learning}}.
\newblock {\em Paladyn, Journal of Behavioral Robotics}, 10(1):14--29, Jan.
  2019.

\bibitem{Hafez2020}
M.~B. Hafez, C.~Weber, M.~Kerzel, and S.~Wermter.
\newblock Improving {{Robot Dual}}-{{System Motor Learning}} with
  {{Intrinsically Motivated Meta}}-{{Control}} and {{Latent}}-{{Space
  Experience Imagination}}.
\newblock {\em Robotics and Autonomous Systems}, 133:103630, Nov. 2020.

\bibitem{Hassan2018}
H.~Hassan, A.~Aue, C.~Chen, V.~Chowdhary, J.~Clark, C.~Federmann, X.~Huang,
  M.~{Junczys-Dowmunt}, W.~Lewis, M.~Li, S.~Liu, T.-Y. Liu, R.~Luo, A.~Menezes,
  T.~Qin, F.~Seide, X.~Tan, F.~Tian, L.~Wu, S.~Wu, Y.~Xia, D.~Zhang, Z.~Zhang,
  and M.~Zhou.
\newblock Achieving {{Human Parity}} on {{Automatic Chinese}} to {{English News
  Translation}}.
\newblock {\em arXiv:1803.05567 [cs]}, June 2018.

\bibitem{NIPS2017}
D.~He, H.~Lu, Y.~Xia, T.~Qin, L.~Wang, and T.-Y. Liu.
\newblock Decoding with {{Value Networks}} for {{Neural Machine Translation}}.
\newblock In {\em International {{Conference}} on {{Neural Information
  Processing Systems}} ({{NIPS}})}, volume 30th, pages 177--186, {Long Beach,
  CA, USA}, Dec. 2017. {Curran Associates Inc.}

\bibitem{He2016a}
D.~He, Y.~Xia, T.~Qin, L.~Wang, N.~Yu, T.-Y. Liu, and W.-Y. Ma.
\newblock Dual {{Learning}} for {{Machine Translation}}.
\newblock In {\em Advances in {{Neural Information Processing Systems}}
  ({{NIPS}})}, volume~29, pages 820--828, {Barcelona, Spain}, Dec. 2016.

\bibitem{he2016}
J.~He, J.~Chen, X.~He, J.~Gao, L.~Li, L.~Deng, and M.~Ostendorf.
\newblock Deep {{Reinforcement Learning}} with a {{Natural Language Action
  Space}}.
\newblock In {\em Annual {{Meeting}} of the {{Association}} for {{Computational
  Linguistics}} ({{ACL}})}, volume~54, pages 1621--1630, {Berlin, Germany},
  Aug. 2016. {Association for Computational Linguistics}.

\bibitem{He2017Reinforcement}
J.~He, M.~Ostendorf, and X.~He.
\newblock Reinforcement {{Learning}} with {{External Knowledge}} and
  {{Two-Stage Q-Functions}} for {{Predicting Popular Reddit Threads}}.
\newblock {\em arXiv:1704.06217 [cs]}, Apr. 2017.

\bibitem{He2016Deep}
J.~He, M.~Ostendorf, X.~He, J.~Chen, J.~Gao, L.~Li, and L.~Deng.
\newblock Deep {{Reinforcement Learning}} with a {{Combinatorial Action Space}}
  for {{Predicting Popular Reddit Threads}}.
\newblock In {\em Conference on {{Empirical Methods}} in {{Natural Language
  Processing}} ({{EMNLP}})}, pages 1838--1848, {Austin, TX, USA}, Nov. 2016.
  {Association for Computational Linguistics}.

\bibitem{heinrich2020}
S.~Heinrich, Y.~Yao, T.~Hinz, Z.~Liu, T.~Hummel, M.~Kerzel, C.~Weber, and
  S.~Wermter.
\newblock Crossmodal {{Language Grounding}} in an {{Embodied Neurocognitive
  Model}}.
\newblock {\em Frontiers in Neurorobotics}, 14, 2020.

\bibitem{henderson2008}
J.~Henderson, O.~Lemon, and K.~Georgila.
\newblock Hybrid {{Reinforcement}}/{{Supervised Learning}} of {{Dialogue
  Policies}} from {{Fixed Data Sets}}.
\newblock {\em Computational Linguistics}, 34(4):487--511, July 2008.

\bibitem{higashinaka2015}
R.~Higashinaka, M.~Mizukami, K.~Funakoshi, M.~Araki, H.~Tsukahara, and
  Y.~Kobayashi.
\newblock Fatal or {{Not}}? {{Finding Errors That Lead}} to {{Dialogue
  Breakdowns}} in {{Chat}}-{{Oriented Dialogue Systems}}.
\newblock In {\em Conference on {{Empirical Methods}} in {{Natural Language
  Processing}} ({{EMNLP}})}, pages 2243--2248, {Lisbon, Portugal}, Sept. 2015.
  {Association for Computational Linguistics}.

\bibitem{Hochreiter1997}
S.~Hochreiter and J.~Schmidhuber.
\newblock Long {{Short}}-{{Term Memory}}.
\newblock {\em Neural Computation}, 9(8):1735--1780, Nov. 1997.

\bibitem{Hutchins1992}
W.~J. Hutchins and H.~L. Somers.
\newblock {\em An {{Introduction}} to {{Machine Translation}}}.
\newblock {Academic Press}, {London}, Apr. 1992.

\bibitem{jiang2012}
J.~Jiang, A.~Teichert, J.~Eisner, and H.~Daum{\'e}~III.
\newblock Learned {{Prioritization}} for {{Trading Off Accuracy}} and
  {{Speed}}.
\newblock In {\em Advances in {{Neural Information Processing Systems}}
  ({{NIPS}})}, volume~25, {Lake Tahoe, NV, USA}, Dec. 2012.

\bibitem{jurcicek2010}
F.~Jurcicek, B.~Thomson, S.~Keizer, F.~Mairesse, M.~Gasic, K.~Yu, and S.~J.
  Young.
\newblock Natural {{Belief}}-{{Critic}}: {{A Reinforcement Algorithm}} for
  {{Parameter Estimation}} in {{Statistical Spoken Dialogue Systems}}.
\newblock In {\em Annual {{Conference}} of the {{International Speech
  Communication Association}} ({{INTERSPEECH}})}, pages 90--93, {Makuhari,
  Japan}, Sept. 2010.

\bibitem{Kalchbrenner2013}
N.~Kalchbrenner and P.~Blunsom.
\newblock Recurrent {{Continuous Translation Models}}.
\newblock In {\em Conference on {{Empirical Methods}} in {{Natural Language
  Processing}} ({{EMNLP}})}, pages 1700--1709, {Seattle, WA, USA}, Oct. 2013.
  {Association for Computational Linguistics}.

\bibitem{Keneshloo2019}
Y.~Keneshloo, T.~Shi, N.~Ramakrishnan, and C.~K. Reddy.
\newblock Deep {{Reinforcement Learning}} for {{Sequence}}-to-{{Sequence
  Models}}.
\newblock {\em IEEE Transactions on Neural Networks and Learning Systems},
  31(7):2469--2489, July 2020.

\bibitem{Kiros2015}
R.~Kiros, Y.~Zhu, R.~R. Salakhutdinov, R.~Zemel, R.~Urtasun, A.~Torralba, and
  S.~Fidler.
\newblock Skip-{{Thought Vectors}}.
\newblock In {\em Advances in {{Neural Information Processing Systems}}
  ({{NIPS}})}, volume~28, pages 3294--3302, {Montreal, QC, Canada}, 2015.
  {Curran Associates, Inc.}

\bibitem{Koehn2010}
P.~Koehn.
\newblock {\em Statistical {{Machine Translation}}}.
\newblock {Cambridge University Press}, {Cambridge ; New York}, Dec. 2009.

\bibitem{Koehn2003}
P.~Koehn, F.~J. Och, and D.~Marcu.
\newblock Statistical {{Phrase}}-{{Based Translation}}.
\newblock In {\em Conference of the {{North American Chapter}} of the
  {{Association}} for {{Computational Linguistics}} on {{Human Language
  Technology}} (HLT-NAACL)}, pages 48--54, {Edmonton, AB, Canada}, May 2003.
  {Association for Computational Linguistics}.

\bibitem{Kubler2009}
S.~K{\"u}bler, R.~McDonald, and J.~Nivre.
\newblock Dependency {{Parsing}}.
\newblock {\em Synthesis Lectures on Human Language Technologies}, 2(1):1--127,
  Dec. 2008.

\bibitem{Kudashkina2020}
K.~Kudashkina, P.~M. Pilarski, and R.~S. Sutton.
\newblock Document-{{Editing Assistants}} and {{Model}}-{{Based Reinforcement
  Learning}} as a {{Path}} to {{Conversational AI}}.
\newblock {\em arXiv:2008.12095 [cs]}, Aug. 2020.

\bibitem{Lam2019}
T.~K. Lam, S.~Schamoni, and S.~Riezler.
\newblock Interactive-{{Predictive Neural Machine Translation Through
  Reinforcement}} and {{Imitation}}.
\newblock In {\em Proceedings of {{Machine Translation Summit XVII Volume}} 1:
  {{Research Track}}}, pages 96--106, {Dublin, Ireland}, Aug. 2019. {European
  Association for Machine Translation}.

\bibitem{Langford2007}
J.~Langford and T.~Zhang.
\newblock The {{Epoch}}-{{Greedy Algorithm}} for {{Contextual Multi}}-armed
  {{Bandits}}.
\newblock In {\em Advances in {{Neural Information Processing Systems}}
  ({{NIPS}})}, volume 20th, pages 817--824, {Vancouver, BC, Canada}, 2007.
  {Curran Associates Inc.}

\bibitem{le2017}
M.~L{\^e} and A.~Fokkens.
\newblock Tackling {{Error Propagation Through Reinforcement Learning}}: {{A
  Case}} of {{Greedy Dependency Parsing}}.
\newblock In {\em Conference of the {{European Chapter}} of the {{Association}}
  for {{Computational Linguistics}} ({{EACL}})}, volume~1, pages 677--687,
  {Valencia, Spain}, Apr. 2017. {Association for Computational Linguistics}.

\bibitem{LeMikolov2014}
Q.~Le and T.~Mikolov.
\newblock Distributed {{Representations}} of {{Sentences}} and {{Documents}}.
\newblock In {\em International {{Conference}} on {{Machine Learning}}
  ({{ICML}})}, volume 32nd, pages 1188--1196, {Beijing, China}, June 2014.
  {PMLR}.

\bibitem{LeCun2015Deepa}
Y.~LeCun, Y.~Bengio, and G.~Hinton.
\newblock Deep {{Learning}}.
\newblock {\em Nature}, 521(7553):436--444, May 2015.

\bibitem{lemon2011}
O.~Lemon.
\newblock Learning {{What}} to {{Say}} and {{How}} to {{Say It}}: {{Joint
  Optimisation}} of {{Spoken Dialogue Management}} and {{Natural Language
  Generation}}.
\newblock {\em Computer Speech \& Language}, 25(2):210--221, Apr. 2011.

\bibitem{levin2000}
E.~Levin, R.~Pieraccini, and W.~Eckert.
\newblock A {{Stochastic Model}} of {{Human}}-{{Machine Interaction}} for
  {{Learning Dialog Strategies}}.
\newblock {\em IEEE Transactions on Speech and Audio Processing}, 8(1):11--23,
  Jan. 2000.

\bibitem{JiweiLi2016}
J.~Li, W.~Monroe, A.~Ritter, M.~Galley, J.~Gao, and D.~Jurafsky.
\newblock Deep {{Reinforcement Learning}} for {{Dialogue Generation}}.
\newblock In {\em Conference on {{Empirical Methods}} in {{Natural Language
  Processing}} (EMNLP)}, pages 1192--1202, {Austin, TX, USA}, Nov. 2016.
  {Association for Computational Linguistics}.

\bibitem{Li2010}
L.~Li, W.~Chu, J.~Langford, and R.~E. Schapire.
\newblock A {{Contextual}}-{{Bandit Approach}} to {{Personalized News Article
  Recommendation}}.
\newblock In {\em International {{Conference}} on {{World Wide Web}}
  ({{WWW}})}, volume 19th, pages 661--670, {Raleigh, NC, USA}, Apr. 2010.
  {Association for Computing Machinery}.

\bibitem{liX2017}
X.~Li, Y.-N. Chen, L.~Li, J.~Gao, and A.~Celikyilmaz.
\newblock End-to-{{End Task}}-{{Completion Neural Dialogue Systems}}.
\newblock In {\em International {{Joint Conference}} on {{Natural Language
  Processing}} (IJCNLP)}, pages 733--743, {Taipei, Taiwan}, Nov. 2017. {Asian
  Federation of Natural Language Processing}.

\bibitem{liX2016}
X.~Li, Z.~C. Lipton, B.~Dhingra, L.~Li, J.~Gao, and Y.-N. Chen.
\newblock A {{User Simulator}} for {{Task}}-{{Completion Dialogues}}.
\newblock {\em arXiv:1612.05688 [cs]}, Nov. 2017.

\bibitem{Li2018Paraphrase}
Z.~Li, X.~Jiang, L.~Shang, and H.~Li.
\newblock Paraphrase {{Generation}} with {{Deep Reinforcement Learning}}.
\newblock In {\em Conference on {{Empirical Methods}} in {{Natural Language
  Processing}} ({{EMNLP}})}, pages 3865--3878, {Brussels, Belgium}, 2018.
  {Association for Computational Linguistics}.

\bibitem{Lillicrap2015}
T.~P. Lillicrap, J.~J. Hunt, A.~Pritzel, N.~Heess, T.~Erez, Y.~Tassa,
  D.~Silver, and D.~Wierstra.
\newblock Continuous {{Control}} with {{Deep Reinforcement Learning}}.
\newblock {\em arXiv:1509.02971}, Sept. 2015.

\bibitem{Lin2017}
K.~Lin, D.~Li, X.~He, Z.~Zhang, and M.-t. Sun.
\newblock Adversarial {{Ranking}} for {{Language Generation}}.
\newblock In {\em Advances in {{Neural Information Processing Systems}}
  ({{NIPS}})}, volume~30, {Long Beach, CA, USA}, Dec. 2017. {Curran Associates,
  Inc.}

\bibitem{litman2000}
D.~J. Litman, M.~S. Kearns, S.~P. Singh, and M.~A. Walker.
\newblock Automatic {{Optimization}} of {{Dialogue Management}}.
\newblock In {\em International {{Conference}} on {{Computational Linguistics}}
  ({{COLING}})}, volume 18th, pages 502--508, {Saarbr\"ucken, Germany}, Aug.
  2000. {Association for Computational Linguistics}.

\bibitem{Liu2020}
Q.~Liu, Y.~Chen, B.~Chen, J.-G. Lou, Z.~Chen, B.~Zhou, and D.~Zhang.
\newblock You {{Impress Me}}: {{Dialogue Generation Via Mutual Persona
  Perception}}.
\newblock In {\em Annual {{Meeting}} of the {{Association}} for {{Computational
  Linguistics}} ({{ACL}})}, volume 58th, pages 1417--1427, {Online}, July 2020.
  {Association for Computational Linguistics}.

\bibitem{Lu2019}
K.~Lu, S.~Zhang, and X.~Chen.
\newblock Goal-{{Oriented Dialogue Policy Learning}} from {{Failures}}.
\newblock {\em Proceedings of the AAAI Conference on Artificial Intelligence},
  33(01):2596--2603, July 2019.

\bibitem{Luketina2019ASO}
J.~Luketina, N.~Nardelli, G.~Farquhar, J.~Foerster, J.~Andreas,
  E.~Grefenstette, S.~Whiteson, and T.~Rockt{\"a}schel.
\newblock A {{Survey}} of {{Reinforcement Learning Informed}} by {{Natural
  Language}}.
\newblock In {\em International {{Joint Conference}} on {{Artificial
  Intelligence}} ({{IJCAI}})}, 28th, pages 6309--6317, {Macau, China}, Aug.
  2019.

\bibitem{mesgar-etal-2021-improving}
M.~Mesgar, E.~Simpson, and I.~Gurevych.
\newblock Improving {{Factual Consistency Between}} a {{Response}} and
  {{Persona Facts}}.
\newblock In {\em Conference of the {{European Chapter}} of the {{Association}}
  for {{Computational Linguistics}} ({{EACL}})}, volume Main Volume, pages
  549--562, {Online}, 2021. {Association for Computational Linguistics}.

\bibitem{Mikolov2013}
T.~Mikolov, K.~Chen, G.~Corrado, and J.~Dean.
\newblock Efficient {{Estimation}} of {{Word Representations}} in {{Vector
  Space}}.
\newblock {\em arXiv:1301.3781 [cs]}, Sept. 2013.

\bibitem{mnih2016}
V.~Mnih, A.~P. Badia, M.~Mirza, A.~Graves, T.~Lillicrap, T.~Harley, D.~Silver,
  and K.~Kavukcuoglu.
\newblock Asynchronous {{Methods}} for {{Deep Reinforcement Learning}}.
\newblock In {\em International {{Conference}} on {{Machine Learning}}
  ({{ICML}})}, volume~48 of {\em 33}, pages 1928--1937, {New York, NY, USA},
  June 2016. {Proceedings of Machine Learning Research (PMLR)}.

\bibitem{Mnih2015}
V.~Mnih, K.~Kavukcuoglu, D.~Silver, A.~A. Rusu, J.~Veness, M.~G. Bellemare,
  A.~Graves, M.~Riedmiller, A.~K. Fidjeland, G.~Ostrovski, S.~Petersen,
  C.~Beattie, A.~Sadik, I.~Antonoglou, H.~King, D.~Kumaran, D.~Wierstra,
  S.~Legg, and D.~Hassabis.
\newblock Human-{{Level Control Through Deep Reinforcement Learning}}.
\newblock {\em Nature}, 518(7540):529--533, Feb. 2015.

\bibitem{Mordatch2018}
I.~Mordatch and P.~Abbeel.
\newblock Emergence of {{Grounded Compositional Language}} in {{Multi}}-{{Agent
  Populations}}.
\newblock {\em Proceedings of the AAAI Conference on Artificial Intelligence},
  32(1), Apr. 2018.

\bibitem{Narasimhan2018}
K.~Narasimhan, R.~Barzilay, and T.~Jaakkola.
\newblock Grounding {{Language}} for {{Transfer}} in {{Deep Reinforcement
  Learning}}.
\newblock {\em Journal of Artificial Intelligence Research}, 63:849--874, Dec.
  2018.

\bibitem{narasimhan2015}
K.~Narasimhan, T.~D. Kulkarni, and R.~Barzilay.
\newblock Language {{Understanding}} for {{Text}}-{{Based Games Using Deep
  Reinforcement Learning}}.
\newblock In {\em Conference on {{Empirical Methods}} for {{Natural Language
  Processing}} ({{EMNLP}})}, pages 1--11, {Lisbon, Portugal}, Sept. 2015.
  {Association for Computational Linguistics}.

\bibitem{Narasimhan2016Improving}
K.~Narasimhan, A.~Yala, and R.~Barzilay.
\newblock Improving {{Information Extraction}} by {{Acquiring External
  Evidence}} with {{Reinforcement Learning}}.
\newblock In {\em Conference on {{Empirical Methods}} in {{Natural Language
  Processing}} ({{EMNLP}})}, pages 2355--2365, {Austin, TX, USA}, Nov. 2016.
  {Association for Computational Linguistics}.

\bibitem{neu2009}
G.~Neu and C.~Szepesv{\'a}ri.
\newblock Training {{Parsers}} by {{Inverse Reinforcement Learning}}.
\newblock {\em Machine Learning}, 77(2):303, Apr. 2009.

\bibitem{Ng2000}
A.~Y. Ng and S.~J. Russell.
\newblock Algorithms for {{Inverse Reinforcement Learning}}.
\newblock In {\em International {{Conference}} on {{Machine Learning}}
  ({{ICML}})}, volume 17th, pages 663--670, {Stanford, CA, USA}, June 2000.
  {Morgan Kaufmann Publishers Inc.}

\bibitem{Och2003}
F.~J. Och.
\newblock Minimum {{Error Rate Training}} in {{Statistical Machine
  Translation}}.
\newblock In {\em Annual {{Meeting}} on {{Association}} for {{Computational
  Linguistics}} ({{ACL}})}, volume~1 of {\em 41st}, pages 160--167, {Sapporo,
  Japan}, July 2003. {Association for Computational Linguistics}.

\bibitem{papaioannou2017}
I.~Papaioannou and O.~Lemon.
\newblock Combining {{Chat}} and {{Task}}-{{Based Multimodal Dialogue}} for
  {{More Engaging HRI}}: {{A Scalable Method Using Reinforcement Learning}}.
\newblock In {\em {{ACM}}/{{IEEE International Conference}} on
  {{Human}}-{{Robot Interaction}} ({{HRI}})}, pages 365--366, {Vienna,
  Austria}, Mar. 2017. {ACM}.

\bibitem{Plato2020}
A.~Papangelis, M.~Namazifar, C.~Khatri, Y.-C. Wang, P.~Molino, and G.~Tur.
\newblock Plato {{Dialogue System}}: {{A Flexible Conversational AI Research
  Platform}}.
\newblock {\em arXiv:2001.06463 [cs]}, Jan. 2020.

\bibitem{Plato2019}
A.~Papangelis, Y.-C. Wang, P.~Molino, and G.~Tur.
\newblock Collaborative {{Multi}}-{{Agent Dialogue Model Training Via
  Reinforcement Learning}}.
\newblock In {\em Annual {{SIGdial Meeting}} on {{Discourse}} and {{Dialogue}}
  ({{SIGDIAL}})}, volume 20th, pages 92--102, {Stockholm, Sweden}, Sept. 2019.
  {Association for Computational Linguistics}.

\bibitem{Papineni2002}
K.~Papineni, S.~Roukos, T.~Ward, and W.-J. Zhu.
\newblock Bleu: {{A Method}} for {{Automatic Evaluation}} of {{Machine
  Translation}}.
\newblock In {\em Annual {{Meeting}} of the {{Association}} for {{Computational
  Linguistics}} ({{ACL}})}, volume 40th, pages 311--318, {Philadelphia,
  Pennsylvania, USA}, July 2002. {Association for Computational Linguistics}.

\bibitem{Pennington2014}
J.~Pennington, R.~Socher, and C.~Manning.
\newblock {{GloVe}}: {{Global Vectors}} for {{Word Representation}}.
\newblock In {\em Conference on {{Empirical Methods}} in {{Natural Language
  Processing}} ({{EMNLP}})}, pages 1532--1543, {Doha, Qatar}, 2014.
  {Association for Computational Linguistics}.

\bibitem{Peters2018}
M.~Peters, M.~Neumann, M.~Iyyer, M.~Gardner, C.~Clark, K.~Lee, and
  L.~Zettlemoyer.
\newblock Deep {{Contextualized Word Representations}}.
\newblock In {\em Conference of the {{North American Chapter}} of the
  {{Association}} for {{Computational Linguistics}}: {{Human Language
  Technologies}} (NAACL-HLT)}, pages 2227--2237, {New Orleans, LA, USA}, June
  2018. {Association for Computational Linguistics}.

\bibitem{Poljak1973}
B.~T. Poljak.
\newblock Pseudogradient {{Adaptation}} and {{Training Algorithms}}.
\newblock {\em Avtomatika i Telemehanika}, 3:45--68, 1973.

\bibitem{RENW20}
F.~R{\"o}der, M.~Eppe, P.~D.~H. Nguyen, and S.~Wermter.
\newblock Curious {{Hierarchical Actor}}-{{Critic Reinforcement Learning}}.
\newblock In {\em International {{Conference}} on {{Artificial Neural
  Networks}} ({{ICANN}})}, Lecture {{Notes}} in {{Computer Science}}, pages
  408--419, {Bratislava, Slovakia}, Sept. 2020. {Springer International
  Publishing}.

\bibitem{Ruckle2018}
A.~R{\"u}ckl{\'e}, S.~Eger, M.~Peyrard, and I.~Gurevych.
\newblock Concatenated {{Power Mean Word Embeddings}} as {{Universal
  Cross}}-{{Lingual Sentence Representations}}.
\newblock {\em arXiv:1803.01400 [cs]}, Sept. 2018.

\bibitem{Russell2009}
S.~Russell and P.~Norvig.
\newblock {\em Artificial {{Intelligence}}: {{A Modern Approach}}}.
\newblock {Pearson}, {Harlow}, 3rd edition, 2010.

\bibitem{Sankar2019}
C.~Sankar and S.~Ravi.
\newblock Deep {{Reinforcement Learning}} for {{Modeling Chit}}-{{Chat Dialog}}
  with {{Discrete Attributes}}.
\newblock In {\em Annual {{SIGdial Meeting}} on {{Discourse}} and
  {{Dialogue}}}, volume 20th, pages 1--10, {Stockholm, Sweden}, Sept. 2019.
  {Association for Computational Linguistics}.

\bibitem{Schatzmann2009Hidden}
J.~Schatzmann and S.~Young.
\newblock The {{Hidden Agenda User Simulation Model}}.
\newblock {\em IEEE Transactions on Audio, Speech, and Language Processing},
  17(4):733--747, May 2009.

\bibitem{Schrittwieser2019}
J.~Schrittwieser, I.~Antonoglou, T.~Hubert, K.~Simonyan, L.~Sifre, S.~Schmitt,
  A.~Guez, E.~Lockhart, D.~Hassabis, T.~Graepel, T.~Lillicrap, and D.~Silver.
\newblock Mastering {{Atari}}, {{Go}}, {{Chess}} and {{Shogi}} by {{Planning}}
  with a {{Learned Model}}.
\newblock {\em Nature}, 588(7839):604--609, Dec. 2020.

\bibitem{schulman2015}
J.~Schulman, S.~Levine, P.~Abbeel, M.~Jordan, and P.~Moritz.
\newblock Trust {{Region Policy Optimization}}.
\newblock In {\em International {{Conference}} on {{Machine Learning}}
  ({{ICML}})}, volume~37, pages 1889--1897, {Lille, France}, July 2015.
  {Proceedings of Machine Learning Research (PMLR)}.

\bibitem{Serban2015}
I.~V. Serban, R.~Lowe, P.~Henderson, L.~Charlin, and J.~Pineau.
\newblock A {{Survey}} of {{Available Corpora For Building Data}}-{{Driven
  Dialogue Systems}}: {{The Journal Version}}.
\newblock {\em Dialogue \& Discourse}, 9(1):1--49, May 2018.

\bibitem{Shi2018}
Z.~Shi, X.~Chen, X.~Qiu, and X.~Huang.
\newblock Toward {{Diverse Text Generation}} with {{Inverse Reinforcement
  Learning}}.
\newblock In {\em International {{Joint Conference}} on {{Artificial
  Intelligence}} ({{IJCAI}})}, volume 27th, pages 4361--4367, {Stockholm,
  Sweden}, July 2018.

\bibitem{Shum2018}
H.-y. Shum, X.-d. He, and D.~Li.
\newblock From {{Eliza}} to {{XiaoIce}}: {{Challenges}} and {{Opportunities}}
  with {{Social Chatbots}}.
\newblock {\em Frontiers of Information Technology \& Electronic Engineering},
  19(1):10--26, Jan. 2018.

\bibitem{Silver2016}
D.~Silver, A.~Huang, C.~J. Maddison, A.~Guez, L.~Sifre, G.~{van den Driessche},
  J.~Schrittwieser, I.~Antonoglou, V.~Panneershelvam, M.~Lanctot, S.~Dieleman,
  D.~Grewe, J.~Nham, N.~Kalchbrenner, I.~Sutskever, T.~Lillicrap, M.~Leach,
  K.~Kavukcuoglu, T.~Graepel, and D.~Hassabis.
\newblock Mastering the {{Game}} of {{Go}} with {{Deep Neural Networks}} and
  {{Tree Search}}.
\newblock {\em Nature}, 529(7587):484--489, Jan. 2016.

\bibitem{Silver2014}
D.~Silver, G.~Lever, N.~Heess, T.~Degris, D.~Wierstra, and M.~Riedmiller.
\newblock Deterministic {{Policy Gradient Algorithms}}.
\newblock In {\em International {{Conference}} on {{Machine Learning}}
  ({{ICML}})}, volume~32 of {\em 31}, pages 387--395, {Beijing, China}, 2014.
  {Proceedings of Machine Learning Research (PMLR)}.

\bibitem{silver2017}
D.~Silver, J.~Schrittwieser, K.~Simonyan, I.~Antonoglou, A.~Huang, A.~Guez,
  T.~Hubert, L.~Baker, M.~Lai, A.~Bolton, Y.~Chen, T.~Lillicrap, F.~Hui,
  L.~Sifre, G.~{van den Driessche}, T.~Graepel, and D.~Hassabis.
\newblock Mastering the {{Game}} of {{Go Without Human Knowledge}}.
\newblock {\em Nature}, 550(7676):354--359, Oct. 2017.

\bibitem{singh2000}
S.~Singh, M.~Kearns, D.~J. Litman, and M.~A. Walker.
\newblock Empirical {{Evaluation}} of a {{Reinforcement Learning Spoken
  Dialogue System}}.
\newblock In {\em National {{Conference}} on {{Artificial Intelligence}}
  ({{AAAI}})}, volume~17, pages 645--651, {Austin, TX, USA}, Aug. 2000. {AAAI
  Press}.

\bibitem{singh2002}
S.~P. Singh, D.~Litman, M.~Kearns, and M.~Walker.
\newblock Optimizing {{Dialogue Management}} with {{Reinforcement Learning}}:
  {{Experiments}} with the {{NJFun System}}.
\newblock {\em Journal of Artificial Intelligence Research}, 16:105--133, Feb.
  2002.

\bibitem{Sipser2013}
M.~Sipser.
\newblock {\em Introduction to the {{Theory}} of {{Computation}}}.
\newblock {Course Technology Cengage Learning}, {Boston, MA}, 3rd edition,
  2013.

\bibitem{sokolov2016}
A.~Sokolov, J.~Kreutzer, C.~Lo, and S.~Riezler.
\newblock Learning {{Structured Predictors}} from {{Bandit Feedback}} for
  {{Interactive NLP}}.
\newblock In {\em Annual {{Meeting}} of the {{Association}} for {{Computational
  Linguistics}} ({{ACL}})}, volume 54th, pages 1610--1620, {Berlin, Germany},
  Aug. 2016. {Association for Computational Linguistics}.

\bibitem{sokolov2015}
A.~Sokolov, S.~Riezler, and T.~Urvoy.
\newblock Bandit {{Structured Prediction}} for {{Learning}} from {{Partial
  Feedback}} in {{Statistical Machine Translation}}.
\newblock In {\em Proceedings of {{MT Summit XV}}}, pages 160--171, {Miami, FL,
  USA}, Nov. 2015. {Association for Machine Translation in the Americas}.

\bibitem{Stahlberg2020Neural}
F.~Stahlberg.
\newblock Neural {{Machine Translation}}: {{A Review}}.
\newblock {\em Journal of Artificial Intelligence Research}, 69:343--418, Oct.
  2020.

\bibitem{Su2018Reward}
P.-H. Su, M.~Ga{\v s}i{\'c}, and S.~Young.
\newblock Reward {{Estimation}} for {{Dialogue Policy Optimisation}}.
\newblock {\em Computer Speech \& Language}, 51:24--43, Sept. 2018.

\bibitem{Sutskever2014}
I.~Sutskever, O.~Vinyals, and Q.~V. Le.
\newblock Sequence to {{Sequence Learning}} with {{Neural Networks}}.
\newblock In {\em Advances in {{Neural Information Processing Systems}}
  ({{NIPS}})}, volume~27, pages 3104--3112, {Montreal, QC, Canada}, Dec. 2014.
  {Curran Associates, Inc.}

\bibitem{Sutton2018}
R.~S. Sutton and A.~G. Barto.
\newblock {\em Reinforcement {{Learning}}: {{An Introduction}}}.
\newblock Adaptive Computation and Machine Learning Series. {The MIT Press},
  {Cambridge, MA}, 2nd edition, Nov. 2018.

\bibitem{tamar2016}
A.~Tamar, Y.~WU, G.~Thomas, S.~Levine, and P.~Abbeel.
\newblock Value {{Iteration Networks}}.
\newblock In {\em Advances in {{Neural Information Processing Systems}}
  ({{NIPS}})}, volume~29, pages 2154--2162, {Barcelona, Spain}, Dec. 2016.
  {Curran Associates, Inc.}

\bibitem{tan2020embodied}
S.~Tan and H.~Liu.
\newblock Towards {{Embodied Scene Description}}.
\newblock In {\em Robotics: {{Science}} and {{Systems}}}, {Corvalis, OR, USA},
  2020. {RSS Foundation}.

\bibitem{thomson2010}
B.~Thomson and S.~Young.
\newblock Bayesian {{Update}} of {{Dialogue State}}: {{A POMDP Framework}} for
  {{Spoken Dialogue Systems}}.
\newblock {\em Computer Speech \& Language}, 24(4):562--588, Oct. 2010.

\bibitem{Ultes2017pydial}
S.~Ultes, L.~M. {Rojas-Barahona}, P.-H. Su, D.~Vandyke, D.~Kim, I.~Casanueva,
  P.~Budzianowski, N.~Mrk{\v s}i{\'c}, T.-H. Wen, M.~Ga{\v s}i{\'c}, and
  S.~Young.
\newblock {{PyDial}}: {{A Multi}}-{{Domain Statistical Dialogue System
  Toolkit}}.
\newblock In {\em Proceedings of {{System Demonstrations}}}, volume 55th, pages
  73--78, {Vancouver, BC, Canada}, July 2017. {Association for Computational
  Linguistics}.

\bibitem{VanHasselt2007}
H.~{van Hasselt} and M.~A. Wiering.
\newblock Reinforcement {{Learning}} in {{Continuous Action Spaces}}.
\newblock In {\em {{IEEE Symposium}} on {{Approximate Dynamic Programming}} and
  {{Reinforcement Learning}} ({{ADPRL}})}, pages 272--279, {Honolulu, HI, USA},
  Apr. 2007.

\bibitem{vogel2010}
A.~Vogel and D.~Jurafsky.
\newblock Learning to {{Follow Navigational Directions}}.
\newblock In {\em Annual {{Meeting}} of the {{Association}} for {{Computational
  Linguistics}} ({{ACL}})}, volume~48 of {\em {{ACL}}}, pages 806--814,
  {Uppsala, Sweden}, July 2010. {Association for Computational Linguistics}.

\bibitem{walker2000}
M.~A. Walker.
\newblock An {{Application}} of {{Reinforcement Learning}} to {{Dialogue
  Strategy Selection}} in a {{Spoken Dialogue System}} for {{Email}}.
\newblock {\em Journal of Artificial Intelligence Research}, 12:387--416, 2000.

\bibitem{Watkins1989}
C.~J. C.~H. Watkins.
\newblock {\em Learning from {{Delayed Rewards}}}.
\newblock Dissertation, Cambridge University, May 1989.

\bibitem{Way2018}
A.~Way.
\newblock Quality {{Expectations}} of {{Machine Translation}}.
\newblock In J.~Moorkens, S.~Castilho, F.~Gaspari, and S.~Doherty, editors,
  {\em Translation {{Quality Assessment}}: {{From Principles}} to
  {{Practice}}}, volume~1 of {\em Machine {{Translation}}: {{Technologies}} and
  {{Applications}}}, pages 159--178. {Springer International Publishing},
  {Cham}, 2018.

\bibitem{Weaver1949}
W.~Weaver.
\newblock Translation.
\newblock In W.~N. Locke and A.~D. Booth, editors, {\em Machine {{Translation}}
  of {{Languages}}: {{Fourteen Essays}}}, pages 15--23. {The MIT Press},
  {Cambridge, MA}, May 1955.

\bibitem{williams2007}
J.~D. Williams and S.~Young.
\newblock Partially {{Observable Markov Decision Processes}} for {{Spoken
  Dialog Systems}}.
\newblock {\em Computer Speech \& Language}, 21(2):393--422, Apr. 2007.

\bibitem{Williams2016}
P.~Williams, R.~Sennrich, M.~Post, and P.~Koehn.
\newblock {\em Syntax-{{Based Statistical Machine Translation}}}, volume~9 of
  {\em Synthesis {{Lectures}} on {{Human Language Technologies}}}.
\newblock {Morgan \& Claypool Publishers}, Aug. 2016.

\bibitem{Wu2018}
L.~Wu, F.~Tian, T.~Qin, J.~Lai, and T.-Y. Liu.
\newblock A {{Study}} of {{Reinforcement Learning}} for {{Neural Machine
  Translation}}.
\newblock In {\em Conference on {{Empirical Methods}} in {{Natural Language
  Processing}} (EMNLP)}, pages 3612--3621, {Brussels, Belgium}, Oct. 2018.
  {Association for Computational Linguistics}.

\bibitem{Wu2016}
Y.~Wu, M.~Schuster, Z.~Chen, Q.~V. Le, M.~Norouzi, W.~Macherey, M.~Krikun,
  Y.~Cao, Q.~Gao, K.~Macherey, J.~Klingner, A.~Shah, M.~Johnson, X.~Liu,
  {\L}.~Kaiser, S.~Gouws, Y.~Kato, T.~Kudo, H.~Kazawa, K.~Stevens, G.~Kurian,
  N.~Patil, W.~Wang, C.~Young, J.~Smith, J.~Riesa, A.~Rudnick, O.~Vinyals,
  G.~Corrado, M.~Hughes, and J.~Dean.
\newblock Google's {{Neural Machine Translation System}}: {{Bridging}} the
  {{Gap Between Human}} and {{Machine Translation}}.
\newblock {\em Computing Research Repository (CoRR) in arXiv},
  abs/1609.08144:23, 2016.

\bibitem{Wuebker2015}
J.~Wuebker, S.~Muehr, P.~Lehnen, S.~Peitz, and H.~Ney.
\newblock A {{Comparison}} of {{Update Strategies}} for {{Large}}-{{Scale
  Maximum Expected BLEU Training}}.
\newblock In {\em Conference of the {{North American Chapter}} of the
  {{Association}} for {{Computational Linguistics}}: {{Human Language
  Technologies}} ({{NAACL HLT}})}, pages 1516--1526, {Denver, CO, USA}, May
  2015. {Association for Computational Linguistics}.

\bibitem{Xiong2017}
W.~Xiong, T.~Hoang, and W.~Y. Wang.
\newblock {{DeepPath}}: {{A Reinforcement Learning Method}} for {{Knowledge
  Graph Reasoning}}.
\newblock In {\em Conference on {{Empirical Methods}} in {{Natural Language
  Processing}} ({{EMNLP}})}, pages 564--573, {Copenhagen, Denmark}, Sept. 2017.
  {Association for Computational Linguistics}.

\bibitem{Yang2020}
M.~Yang, W.~Huang, W.~Tu, Q.~Qu, Y.~Shen, and K.~Lei.
\newblock Multitask {{Learning}} and {{Reinforcement Learning}} for
  {{Personalized Dialog Generation}}: {{An Empirical Study}}.
\newblock {\em IEEE Transactions on Neural Networks and Learning Systems},
  32(1):49--62, Jan. 2021.

\bibitem{young2010}
S.~Young, M.~Ga{\v s}i{\'c}, S.~Keizer, F.~Mairesse, J.~Schatzmann, B.~Thomson,
  and K.~Yu.
\newblock The {{Hidden Information State Model}}: {{A Practical Framework}} for
  {{POMDP}}-{{Based Spoken Dialogue Management}}.
\newblock {\em Computer Speech \& Language}, 24(2):150--174, Apr. 2010.

\bibitem{young2013}
S.~Young, M.~Ga{\v s}i{\'c}, B.~Thomson, and J.~D. Williams.
\newblock {{POMDP}}-{{Based Statistical Spoken Dialog Systems}}: {{A Review}}.
\newblock {\em Proceedings of the IEEE}, 101(5):1160--1179, May 2013.

\bibitem{young2000}
S.~J. Young.
\newblock Probabilistic {{Methods}} in {{Spoken}}-{{Dialogue Systems}}.
\newblock {\em Philosophical Transactions: Mathematical, Physical and
  Engineering Sciences}, 358(1769):1389--1402, Apr. 2000.

\bibitem{Yu2017-SeqGAN}
L.~Yu, W.~Zhang, J.~Wang, and Y.~Yu.
\newblock {{SeqGAN}}: {{Sequence Generative Adversarial Nets}} with {{Policy
  Gradient}}.
\newblock {\em Proceedings of the AAAI Conference on Artificial Intelligence},
  31(1):2852--2858, Feb. 2017.

\bibitem{yu2017}
Z.~Yu, A.~Rudnicky, and A.~Black.
\newblock Learning {{Conversational Systems}} that {{Interleave Task}} and
  {{Non}}-{{Task Content}}.
\newblock In {\em International {{Joint Conference}} on {{Artificial
  Intelligence}} ({{IJCAI}})}, volume 26th, pages 4214--4220, {Melbourne, VIC,
  Australia}, 2017.

\bibitem{zhang2009}
L.~Zhang and K.~P. Chan.
\newblock Dependency {{Parsing}} with {{Energy}}-{{Based Reinforcement
  Learning}}.
\newblock In {\em International {{Conference}} on {{Parsing Technologies}}
  ({{IWPT}})}, volume 11th, pages 234--237, {Paris, France}, Oct. 2009.
  {Association for Computational Linguistics}.

\bibitem{Zhao2019Rethinking}
T.~Zhao, K.~Xie, and M.~Eskenazi.
\newblock Rethinking {{Action Spaces}} for {{Reinforcement Learning}} in
  {{End}}-to-end {{Dialog Agents}} with {{Latent Variable Models}}.
\newblock In {\em Conference of the {{North American Chapter}} of the
  {{Association}} for {{Computational Linguistics}}: {{Human Language
  Technologies}} ({{NAACL HLT}})}, volume~1, pages 1208--1218, {Minneapolis,
  Minnesota}, June 2019. {Association for Computational Linguistics}.

\bibitem{ZhuCao2020}
S.~Zhu, R.~Cao, and K.~Yu.
\newblock Dual {{Learning}} for {{Semi-Supervised Natural Language
  Understanding}}.
\newblock {\em IEEE/ACM Transactions on Audio, Speech, and Language
  Processing}, 28:1936--1947, 2020.

\bibitem{Ziebart2008}
B.~D. Ziebart, A.~Maas, J.~A. Bagnell, and A.~K. Dey.
\newblock Maximum {{Entropy Inverse Reinforcement Learning}}.
\newblock In {\em National {{Conference}} on {{Artificial Intelligence}}
  ({{AAAI}})}, volume~3 of {\em 23rd}, pages 1433--1438, {Chicago, IL, USA},
  July 2008. {AAAI Press}.

\end{thebibliography}

\end{document}